\documentclass{article}

\PassOptionsToPackage{numbers, sort&compress}{natbib}

    \usepackage[final]{neurips_data_2024}

\usepackage[utf8]{inputenc} 
\usepackage[T1]{fontenc}    
\usepackage{hyperref}       
\usepackage{url}            
\usepackage{booktabs}       
\usepackage{amsfonts}       
\usepackage{nicefrac}       
\usepackage{microtype}      
\usepackage{xcolor}         

\usepackage{graphicx}
\usepackage{caption}
\usepackage{subcaption}

\title{Benchmark Data Repositories for Better Benchmarking}

\author{%
  Rachel Longjohn$^{1}$\thanks{Denotes equal contribution.} , \quad Markelle Kelly$^{2}$\footnotemark[1] , \quad Sameer Singh$^2$,  \quad Padhraic Smyth$^2$ \\
  Department of Statistics$^1$,  Department of Computer Science$^2$\\
  University of California, Irvine\\
  Irvine, CA 92697 \\
  \texttt{\{rlongjoh,kmarke,sameer,pjsmyth\}@uci.edu} \\
  }

\begin{document}

\maketitle

\begin{abstract}

In machine learning research, it is common to evaluate algorithms via their performance on standard benchmark datasets. While a growing body of work establishes guidelines for---and levies criticisms at---data and benchmarking practices in machine learning, comparatively less attention has been paid to the data repositories where these datasets are stored, documented, and shared. In this paper, we 
analyze the landscape of these \textit{benchmark data repositories} and the role they can play in improving benchmarking. This role includes addressing issues with both datasets themselves (e.g., representational harms, construct validity) and the manner in which evaluation is carried out using such datasets (e.g., overemphasis on a few datasets and metrics, lack of reproducibility). 
To this end, we identify and discuss a set of considerations surrounding the design and use of benchmark data repositories, with a focus on improving benchmarking practices in machine learning.

\end{abstract}
\section{Introduction}
Evaluating machine learning (ML) algorithms on benchmark datasets is a central pillar of ML research. This performance benchmarking
facilitates direct comparison across different techniques, which is important, for example, 
in the publication of research that introduces a novel method or for selecting the most appropriate approach for a particular application \citep{Thiyagalingam2022qt, bentejac2021comparative, Hoffmann2019yi, caruana2006empirical, dehghani2021benchmark}. Ideally, these benchmark datasets serve as proxies for real-world tasks, so that performing well on the task represents meaningful advancement 
toward some desired real-world ML capability \citep{raji2021ai, paullada2021data, dotan2020value, denton2020bringing, matters2012wagstaff}. Benchmarking can help quantify progress on these tasks over time, and the availability of a well-studied, standard task evaluation environment can be a critical first step before moving to real-world applications, especially in high-stakes or expensive domains. In addition, evaluating with a benchmark dataset can be useful as a sanity check when developing a new methodology, as well as for ML education and training \cite{hicks2018teaching, serrano2017experiential}.

Early data repositories, such as the UCI ML Repository, arose to address the data needs that come with ML benchmarking \cite{langley2011changing}. These repositories started as relatively small-scale efforts, but as the field of ML has rapidly grown, they have become more sophisticated, supporting additional features such as leaderboards comparing the benchmarked performance of ML models on a given dataset \cite{vanschoren2014openml,bischlopenml,brazdil2022metadata, lhoest2021datasets,banachewicz2022kaggle, dau2019ucr, olson2017pmlb}. ML data repositories are fundamentally different from traditional domain-specific data repositories. For example, they tend to contain datasets from a wide variety of domains, and the process of selecting a dataset is often less about a scientific or engineering application and more about the compositional characteristics of the data and its associated tasks, for which a particular class of methods is applicable, e.g., multivariate spatiotemporal or network datasets.

Today, these ML data repositories---in particular, HuggingFace Datasets, Kaggle, OpenML, Papers with Code Datasets, TensorFlow Datasets, and the UCI ML Repository---are widely used. However, relatively little work has been devoted to understanding them and the specific factors involved in their design. In this paper, we introduce the term \textit{benchmark data repository} to describe repositories that support the discovery and use of datasets for evaluating ML models (for brevity, we will use \textit{benchmark repository} in the remainder of the paper). Our focus is the role of benchmark repositories in ML research, particularly the relationship between benchmark repositories and criticisms of current data and benchmarking practices in ML. \footnote{While in this paper we focus on data used for model evaluation, we note that many of our points are also relevant to pretraining data.}
Each of Sections \ref{sec:datasetsasresearch}-\ref{sec:benchmarkoveruse} reviews one of these issues, progressing through the dataset lifecycle---from creation and development to documentation and sharing to use and reuse for model evaluation \citep{ashmore2021assuring,park2023ripple, koch2021reduced}. For each criticism, we identify ways in which benchmark repositories can be part of the solution,
motivated by existing standards, observed trends in the field, and examples from our experience as repository curators.
We believe that these recommendations will be useful both for the owners of benchmark repositories in designing and improving their repositories, and more broadly to the creators and users of benchmark datasets in determining how to store, document, and find data. To the best of our knowledge, this paper is the first to define and establish best practices specifically for benchmark repositories, as well as to connect the practices of benchmark repositories to ML and data repository practices in general.

\section{Valuing Datasets as Research Contributions}
\label{sec:datasetsasresearch} 
Data work maintains a legacy of being under-valued and under-incentivized by the ML community \cite{gero2023incentive, birhane2022values, sambasivan2021everyone, jo_lessons_2020, hutchinson2021towards, longpre2023pretrainer, bhardwaj2024machine, raji2021ai, heinzerling2020nlp, scheuerman2021datasets}, often regarded as an ``engineering exercise'' \cite{desai2024archival} or ``operational'' \cite{sambasivan2021everyone}. Several recent initiatives, such as the NeurIPS Datasets and Benchmarks track\footnote{\href{https://neuripsconf.medium.com/announcing-the-neurips-2021-datasets-and-benchmarks-track-644e27c1e66c}{https://neuripsconf.medium.com/announcing-the-neurips-2021-datasets-and-benchmarks-track-644e27c1e66c}} and the Journal of Data-centric Machine Learning Research \cite{oala2024dmlr}, have sought to change this pattern by providing peer-reviewed venues for publishing papers on data contributions. 
In this section, we posit that benchmark repositories can also help recognize datasets as intellectual contributions to the scholarly ecosystem by providing
1) dataset citations, 2) ``connection metadata,'' and 3) dataset licenses. Reinforcing the value of data work incentivizes dataset creators to pay greater care and attention during dataset development and documentation---effects that propagate throughout the dataset lifecycle \citep{sambasivan2021everyone, gero2023incentive, jo_lessons_2020}.

\subsection{Dataset Citations and Metrics}
\label{sec:cite-metrics}
For a dataset to operate in the ML ecosystem as a first-class research contribution, researchers must be able to locate it and its metadata via a persistent stable URL (as is the norm with published papers). In particular, the assignment of a \textit{persistent identifier} (PID), such as a DOI, that can reliably be used to access a dataset has been widely recommended by experts \cite{wilkinson2016fair, stodden2016enhancing, goodman2014ten, liangadvances2022, scheuerman2021datasets, whostp2022desirable}. However, ML datasets and their documentation frequently lack PIDs and are often only available via GitHub or personal/research group websites \cite{peng2021mitigating, scheuerman2021datasets, heger2022understanding}. Repositories can help address this by minting DOIs for submitted datasets (e.g., as Kaggle\footnote{\href{https://www.kaggle.com/discussions/product-feedback/108594}{https://www.kaggle.com/discussions/product-feedback/108594}} and HuggingFace Datasets\footnote{\href{https://huggingface.co/blog/introducing-doi}{https://huggingface.co/blog/introducing-doi}} do).

\begin{figure}
    \begin{subfigure}[t]{0.3\textwidth}
         \centering
         \includegraphics[width=\textwidth]{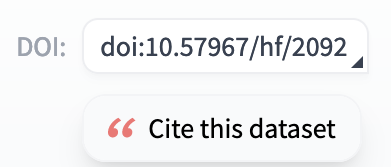}
         \caption{HuggingFace \citep{huggingfacefw_2024}}
     \end{subfigure}
     \hfill
     \begin{subfigure}[t]{0.55\textwidth}
         \centering
         \includegraphics[width=\textwidth]{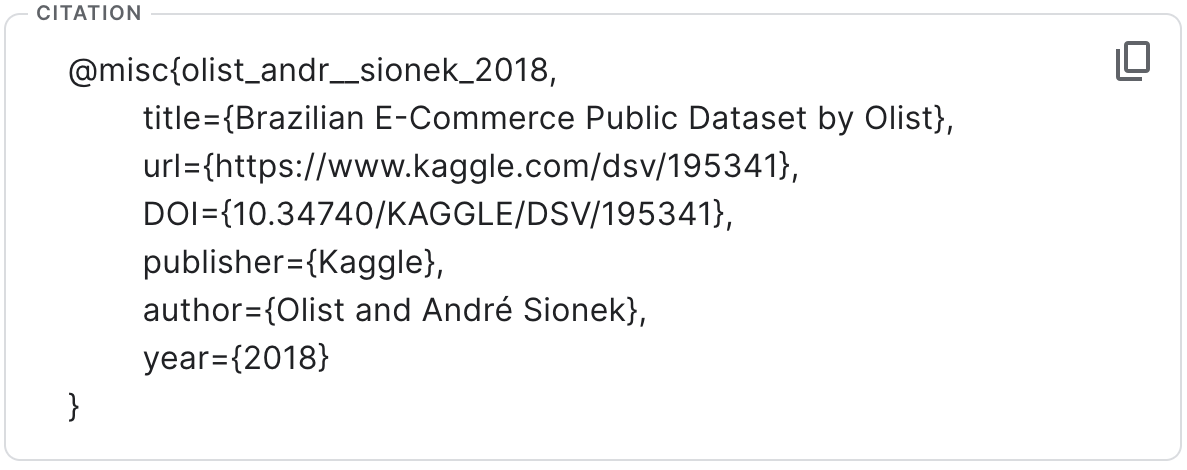}
         \caption{Kaggle \citep{olist_andr__sionek_2018}}
     \end{subfigure}
    \caption{Examples of DOIs and citations in repositories.}
    \label{fig:citation}
\end{figure}

PIDs are the foundation of \textit{dataset citations}, which give proper attribution to dataset creators, rather than solely citing associated publications \cite{stodden2016enhancing, task2013out, jddcp, altman2015introduction, silvello2018theory, groth2020fair, fenner2019data, starr2015achieving, cousijn2018data}. In ML, however, datasets are often referred to using combinations of names, descriptions, and associated papers, which can be challenging to disambiguate \cite{peng2021mitigating}. In contrast, many data repositories already provide standardized dataset citations that can be easily copied in a desired format (e.g., BibTeX) and include the minted DOI (Figure \ref{fig:citation}). Beyond giving credit, citing a dataset enables researchers to track its usage throughout the literature, which is particularly relevant in ML, e.g., for performance comparisons.

Furthermore, metrics such as the number of citations, number of views, or number of downloads can help quantify data impact, highlighting the value of the dataset in terms of its contribution to the ML community and potentially benefiting a variety of stakeholders (e.g., the researchers whose work is being cited or funders assessing a return on investment   \cite{lowenberg_daniella_2019_3525349, Lowenberg2022Recognizing, kratz2015making, fenner2018code}). Repositories can provide the infrastructure for tracking these metrics of interest; for example, OpenML counts the number of times a dataset has been used in experiment runs and the number of times it has been downloaded \citep{vanschoren2014openml}. Data organizations, such as Scholix, the Data Usage Metric Working Group and Project Counter \cite{cousijn2019bringing, counter2018code, scholix2017_1120265}, are working towards more sophisticated frameworks for the provision of data metrics, and benchmark repositories are in a prime position to foster collaborations with these efforts.

\subsection{Connection Metadata}
\label{sec:connectionmetadata}
Repositories can also support the treatment of datasets as research contributions via \textit{connection metadata}, 
which connects a dataset to associated research entities (such as the dataset's creators or maintainers, publications, code, or other datasets) \cite{datacite-connection}.

\begin{figure}
    \begin{subfigure}[t]{0.5\textwidth}
         \centering
         \includegraphics[width=\textwidth]{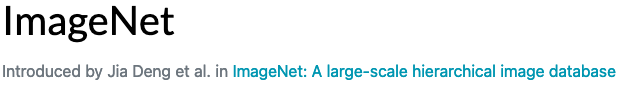}
         \caption{Papers with Code\footnotemark}
     \end{subfigure}
     \hfill
     \begin{subfigure}[t]{0.5\textwidth}
         \centering
         \includegraphics[width=\textwidth]{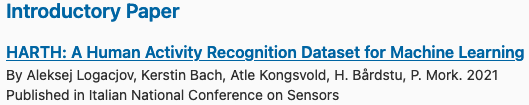}
         \caption{UCI ML Repository \citep{harth_779}}
     \end{subfigure}
    \caption{Examples of connecting datasets to papers in repositories.}
    \label{fig:paper-link}
\end{figure}
\footnotetext{\href{https://paperswithcode.com/dataset/imagenet}{https://paperswithcode.com/dataset/imagenet}}

The dataset construction process is rife with consequential, ``value-laden'' decisions \cite{dotan2020value, longpre2023pretrainer, hutchinson2021towards, desai2024archival}. The rationale behind these decisions may be described in an \textit{introductory paper}: a publication that combines the narrative style of an article with the technical description of a dataset and its design process (also referred to as ``data articles'' \cite{austin2017key} or ``dataset descriptors''\footnote{\href{https://www.nature.com/sdata/journal-information}{https://www.nature.com/sdata/journal-information}}). Introductory papers can give the data a story beyond standardized documentation, providing useful context about the problem, background on data collection procedures, and guidance about tasks for which the data have already been used. When these papers are peer-reviewed, it can lend additional credibility to the dataset for those considering it for re-use. Introductory papers can be included in benchmark repositories as a standardized metadata field (Figure \ref{fig:paper-link}), raising the visibility of these important documents.

Repositories can also identify an individual who agrees to serve as a dataset's \textit{point of contact}: someone responsible for answering questions about the dataset and addressing any issues. 
Ideally, this person is also one of the dataset's creators, as they are best equipped to answer questions about the data and facilitate re-use \cite{pasquetto2019uses, heger2022understanding}. 
Although long-term data maintenance, and determining different stakeholders' responsibilities in that maintenance, remain challenging tasks \cite{borgman2022why, hutchinson2021towards, borgman2019lives}, establishing a point of contact can help prevent the development of a disconnect between a dataset and its creators, which is not uncommon in ML 
\cite{yadav2019cold, radin2017digital, groemping2019south}. By requiring that contact information for a responsible individual 
be specified in metadata, repositories can encourage an ongoing connection between the dataset, those who created it, and those who want to re-use it. 

\subsection{Dataset Licenses}
It is widely recommended 
that datasets come with clear use guidance, often via a \textit{license} \cite{whostp2022desirable, wilkinson2016fair, gebru2021datasheets}. These safeguards can help prevent the unintended use of data, an important part of respecting datasets as intellectual contributions. 
Data repositories can include licenses as part of a dataset's metadata. They can also make selecting a license easier on dataset donors, e.g., by showing which licenses are popular, providing help text and links to the license language, or comparing the salient parts of different popular licenses. For ML datasets in particular, licensing can be complicated (e.g., it is ambiguous if models trained on a dataset count as ``derivative work'' \cite{benjamin2019towards}), and dataset licenses commonly used in other domains may not effectively restrict data use in ML (e.g., training commercial ML pipelines) \cite{peng2021mitigating, rajbahadur2021can}. In addition, current licensing practices for ML datasets are often irregular, conflicting, poorly documented, or over-permissive given the dataset content \citep{duan2024modelgo}; in one survey of over 1800 text datasets, 69\% had ``unspecified'' licenses on HuggingFace \citep{longpre2023data}. To help mitigate these issues, benchmark repositories can encourage the use of licenses that were constructed with ML use cases in mind, such as the Montreal Data License \cite{benjamin2019towards}, or others, as more work is done in this area.

\section{Addressing Issues with Dataset Content}
\label{sec:datasetissues}
Over the past decade or two, numerous issues with common benchmark datasets have been discovered, including technical flaws such as labeling errors and annotation artifacts \citep{northcutt2021pervasive, dehghani2021benchmark, gururangan2018annotation, sasha2021framework}, privacy and copyright violations \citep{raji2021face, bandy_addressing_2021, peng2021mitigating, andrews2024ethical}, inclusions of hate speech or other harmful content \citep{birhane2021large,yang2020towards}, representational biases \citep{buolamwini2018gender, merler2019diversity, park2021understanding, bandy_addressing_2021}, and miscellaneous ethical issues \citep{bostonhousing, radin2017digital, peng2021mitigating}. Without clear documentation or careful data auditing, it is easy for these problems to go undiscovered well after a dataset's initial release and propagate harmful effects to downstream results \citep{andrews2024ethical, liangadvances2022}. Further, even once an issue is discovered, updating or deprecating a dataset can be ineffective \cite{peng2021mitigating, sasha2021framework}. Benchmark repositories can help detect and address dataset issues by collecting contextual metadata, performing quality reviews, and supporting the revision and deprecation of datasets.

\subsection{Contextual Metadata}
\label{sec:contextmetadata}
Benchmark datasets are often disseminated without detailed information about their broader context \citep{raji2021face, hutchinson2021towards, jo_lessons_2020,geiger2020garbage,denton2020bringing, yang2024navigating}. By collecting \textit{contextual metadata}, including information about a dataset's source, funding, collection, annotation, and preprocessing, benchmark repositories can illuminate the assumptions and motivations of dataset creators and flag potential dataset issues \citep{paullada2021data, hand2006classifier, bender2021dangers,raji2021ai, leavy2021ethical, bhardwaj2024machine}.
To this end, several standards and schemata that include contextual metadata have been established, including Data Cards \citep{pushkarna2022data}, datasheets for datasets \citep{gebru2021datasheets}, the Dataset Nutrition Label \citep{holland2020dataset, chmielinski2022dataset}, and the FAIR principles \citep{wilkinson2016fair}. Such metadata can help ML practitioners detect issues earlier \citep{boyd2021datasheets}; several ``retrospective'' datasheets for well-known datasets have demonstrated how contextual information raises red flags and could have contributed to earlier detection of data issues \citep{bandy_addressing_2021, dodge2021documenting}.

In particular, information about the source of a dataset can alert data users to privacy or consent issues, representation biases, the potential for harmful content, or a mismatch with their target domain \citep{biderman2020pitfalls, mehrabi2022survey}. For example, multiple facial image datasets include mugshots or surveillance camera footage \citep{stewart2016end, ristani2016performance, founds2011nist}---raising red flags about the consent and privacy of the photographed individuals. 
Another example is the NIST Face Recognition Vendor Test dataset, which was funded by the U.S. Department of Homeland Security and contains data from the U.S. Mexican visa archive \citep{phillips2003face, dooley2022robustness}.
In the use of this dataset for general facial recognition evaluation (e.g., \citep{zhu2023webface}), its source and original intent are cause for concern about its transferability \citep{raji2021face}.
Generally, understanding the origins of a dataset can help ML researchers determine if it is appropriate for their use case, discouraging the use of benchmarks that are poor proxies for the task they are supposedly evaluating \citep{hand2006classifier,koch2021reduced,paullada2021data,raji2021ai,biderman2020pitfalls}.

Data selection, filtering, and annotation processes are important design decisions that can significantly impact downstream performance \citep{liangadvances2022, andrews2024ethical, rangineni2023analysis, paullada2021data, tsipras2020imagenet}. One example is the systematic exclusion of text authored by or about marginalized groups in large, web-scraped text datasets due to curation and data filtering processes \citep{bandy_addressing_2021,longpre2023pretrainer,desai2024archival,liangadvances2022}. Another pervasive issue is biased annotations, which are often crowd-sourced \cite{shilad2015turkers, bowman2021fix, daneshjou2022disparities, liangadvances2022, sorqvist2011women, jo_lessons_2020}, e.g., as have been documented in the ImageNet dataset \citep{paullada2021data, tsipras2020imagenet}. 
However, these processes tend to be under-documented \citep{longpre2023pretrainer}; for instance, in a survey of over 100 papers introducing computer vision datasets, 36.6\% did not provide any description of the human annotators; only 7.8\% reported annotator demographics \citep{scheuerman2021datasets}.

Clearly documenting data source, intent, collection, and processing procedures sheds light on these dataset issues early on in the data lifecycle. However, dataset creators do not necessarily prioritize metadata on their own
\citep{scheuerman2021datasets, heger2022understanding}, and documentation is often scattered and unstandardized \citep{raji2021face, boyd2021datasheets, geiger2020garbage}. 
Benchmark repositories can work against this pattern by requiring dataset creators to provide detailed documentation of contextual information (and making it easily accessible), e.g., via an accompanying datasheet \citep{gebru2021datasheets} and/or published introductory paper, which thoroughly describe a dataset's context.

\begin{figure}
    \centering
    \includegraphics[width=0.5\textwidth]{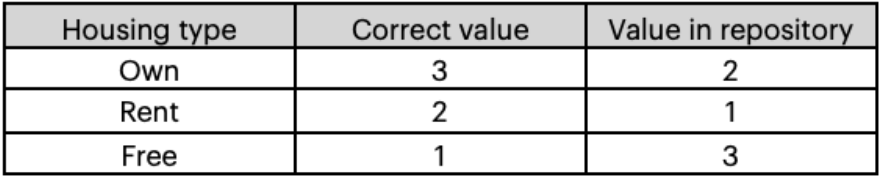}
    \caption{The Statlog (German Credit Data) dataset \cite{statlog_german_credit_data_144},  
    hosted by the UCI ML Repository, is a sample of customer records from  
    a German bank, with the task of classifying each individual as a good or bad credit risk. 
    In the repository documentation, 8
    categorical variables have their levels mixed up or incorrectly described (e.g., see attribute 15, the type of housing the debtor lives in, above). 
    Groemping \cite{groemping2019south} tracked down papers which describe the dataset's origins \cite{haubetaler1979empirische, haubetaler1981methoden, fahrmeir1981kategoriale, fahrmeir1984multivariate} to construct a proper code table. She donated the corrected dataset as the South German Credit dataset in 2019 \cite{groemping2019south} but the original dataset from 1994 has nonetheless been widely used in ML research.}
    \label{fig:german-credit}
\end{figure}

\subsection{Quality Review}
\label{sec:qa}
Ideally, quality issues with benchmark datasets and their metadata are detected early and corrected; otherwise, these concerns should either be documented or used as a rationale to withdraw the dataset. For example, datasets containing personally identifiable information should not be released \citep{jo_lessons_2020} and documentation errors (as in Figure \ref{fig:german-credit}) should be quickly amended. Benchmark repositories can help identify these problems throughout the data lifecycle by (1) performing a pre-release \textit{quality review} \citep{biderman2020pitfalls} to catch issues before a dataset is shared, and (2) by serving as a centralized location to collect users' reports and concerns \citep{sasha2021framework} to flag issues throughout a dataset's use and reuse. 
With stringent quality assurance, ML researchers can reliably look to a repository for high-quality datasets \citep{liangadvances2022}, making it easier to avoid using unvalidated, problematic datasets for benchmarking.

Quality reviews can help counter the current lack of incentive for ML dataset creators to consider ethical issues, which has been pointed to as a major contributor to the numerous ethical problems with benchmark datasets \citep{jo_lessons_2020}. For example, benchmark data collection often does not undergo institutional ethical review \citep{ andrews2024ethical}; in one survey \citep{scheuerman2021datasets}, only 5 out of 100 papers introducing datasets with human subjects mentioned an institutional review board (IRB) or equivalent ethical review. 
As a result there has been a call for more intervention in data curation, involving curators who can focus on developing conduct codes and ethical review processes rather than relying on dataset creators \citep{bender2021dangers, desai2024archival, bhardwaj2024machine, jo_lessons_2020}.
It is an open question to what extent repositories should be involved in these decisions; several popular repositories (e.g., Zenodo, Mendeley) view their role as only providing infrastructure and not conducting any kind of data review. However, we posit that benchmark repositories are well-positioned in the data pipeline to perform at least basic ethical checks and initiate a movement towards interventionism. We point to the growing body of literature on ethical data curation for ML \citep{biderman2020pitfalls, bowman2021fix, leavy2021ethical, andrews2024ethical, srikumar2022advancing, bhardwaj2024machine} as a starting point for the development of ethical review processes.

Conducting thorough quality assurance can be particularly difficult for benchmark repositories because they typically host data from a variety of domains. In contrast, disciplinary repositories, which specialize in a particular domain, often have a community of experts with the knowledge to conduct quality reviews. We point to requiring peer-reviewed introductory papers as a potential step in this direction, as the publishing venue may be able to perform more targeted reviews, and an increasing number of venues also incorporate ethical reviews.\footnote{e.g., \href{https://medium.com/@icml2024pc/ethics-review-at-icml-e3b4ce1afd54}{https://medium.com/@icml2024pc/ethics-review-at-icml-e3b4ce1afd54},\href{https://neurips.cc/public/EthicsGuidelines}{https://neurips.cc/public/EthicsGuidelines}} Repositories could also outsource reviews for datasets via a  network of experts such as the Data Curation Network \footnote{\href{https://datacurationnetwork.org}{https://datacurationnetwork.org}}.

\begin{figure}
    \begin{subfigure}[t]{0.32\textwidth}
         \centering
         \includegraphics[width=\textwidth]{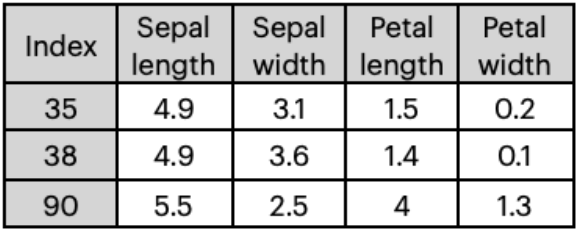}
         \caption{The original data for the 35th, 38th, and 90th iris flowers.}
     \end{subfigure}
     \hfill
     \begin{subfigure}[t]{0.32\textwidth}
         \centering
         \includegraphics[width=\textwidth]{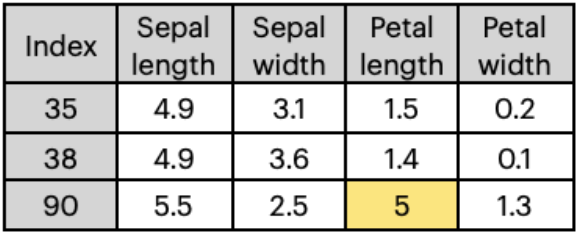}
         \caption{An alternative version of the data with the petal length of the 90th flower incorrect.}
     \end{subfigure}
     \hfill
     \begin{subfigure}[t]{0.32\textwidth}
         \centering
         \includegraphics[width=\textwidth]{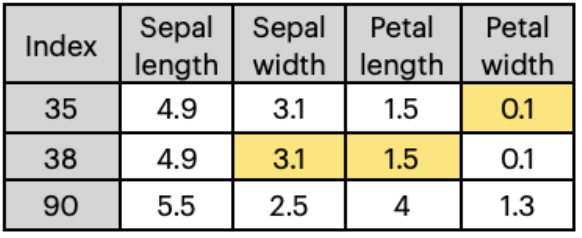}
         \caption{An alternative version of the data with erroneous entries for the 35th and 38th flowers.}
     \end{subfigure}
        \caption{The Iris dataset from the UCI ML Repository is widely used for evaluating clustering and classification algorithms \cite{misc_iris_53}. Each observation corresponds to an iris flower, including sepal and petal measurements and its specific species (out of three classes). After years of use, it was discovered that there were multiple different widely-publicized versions of this dataset, with differing measurements for certain observations. Consequently, the reported performances of classification models on Iris (across a large number of published papers) are not necessarily comparable \cite{bezdek1999will}.}
        \label{fig:iris}
\end{figure}

\subsection{Dataset Revision and Deprecation}
\label{sec:revisiondeprecation}
Although quality review can help catch serious issues before the release of a dataset, inevitably, some datasets will need to be updated, corrected, or deprecated. As a centralized data source, benchmark repositories can help support the revision and deprecation of datasets.

Benchmark repositories can support \textit{dataset revision} by documenting data versions and connecting each dataset to a responsible point of contact. When different versions of a dataset are not clearly associated with unique version numbers, differing versions may be used interchangeably \citep{Sawchuk2021qy, sculley2015hidden} (e.g., as in Figure \ref{fig:iris}). Repositories can enforce versioning by assigning a new version number whenever a data file is changed. Documentation of the revision, including what was changed or removed and a rationale for the changes, should also be provided \citep{biderman2020pitfalls, scheuerman2021datasets}. In addition, by associating datasets with a responsible point of contact (see Section \ref{sec:connectionmetadata}), repositories can help streamline the resolution of questions or issues regarding a dataset.

\begin{figure}
    \centering
    \includegraphics[width=0.8\textwidth]{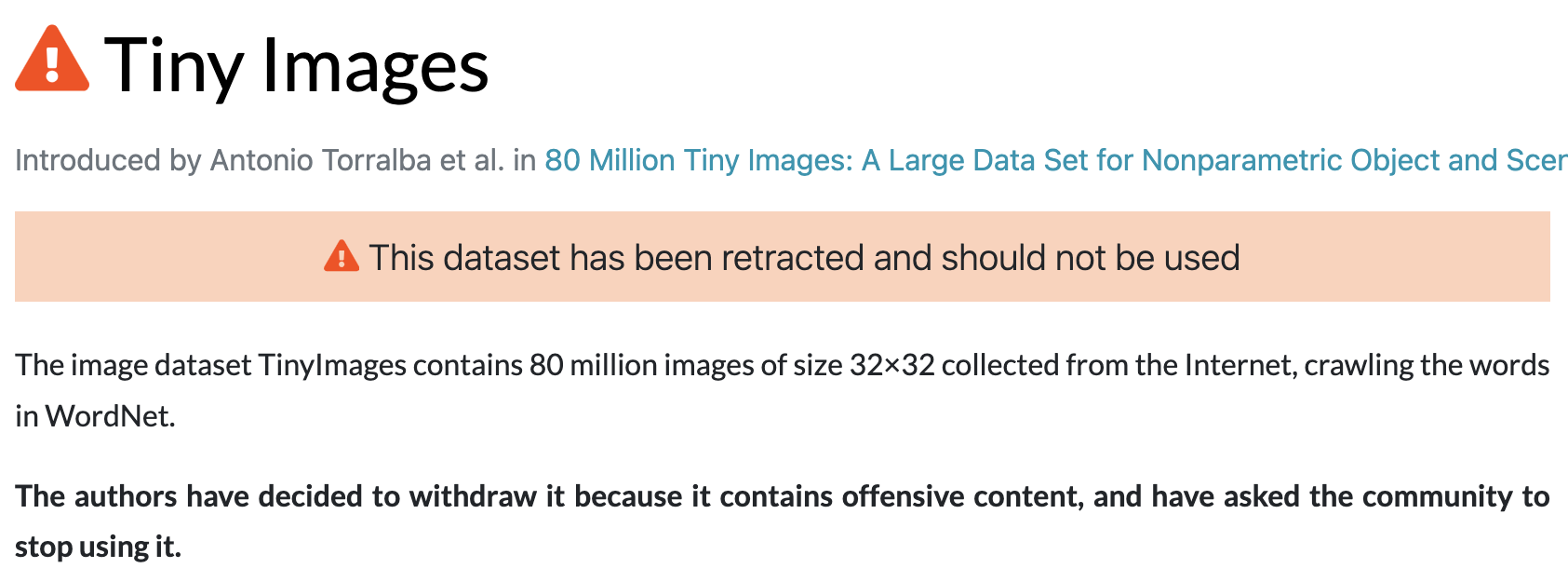}
    \caption{The Papers with Code dataset page for the deprecated Tiny Images dataset.}
    \label{fig:pwc-tiny}
\end{figure}

Repositories can also support the \textit{deprecation} of datasets. Currently, there is no standardized process for dataset deprecation: creators often withdraw their dataset without an explanation of why it was withdrawn or explicit instructions not to use the dataset (or other post-deprecation protocols). For example, in \cite{sasha2021framework}’s case study of six high-profile dataset retractions, three (MS-Celeb-1M, Duke MTMC, and HRT Transgender) did not provide any reason for the dataset's removal. Moreover, deprecation reports are posted in a scattered, decentralized manner via news articles, conference papers, or researcher or lab websites \citep{sasha2021framework}. Ultimately, it can be unclear to researchers if a dataset is acceptable to use; it is not uncommon for datasets to remain in use after their deprecation, including in published, peer-reviewed papers \citep{paullada2021data, peng2021mitigating,sasha2021framework}. To mitigate this, benchmark repositories can (1) establish a process for deprecating a dataset, in which its creators submit a standardized report, detailing the reasons for deprecation and post-deprecation protocols, and (2) maintain a page connected to the dataset DOI (see Section \ref{sec:cite-metrics}) with the deprecation report and original metadata \citep{Sheridan2021xu}. If a deprecation report is clearly displayed in the same place where a dataset was available, it clarifies to researchers (and reviewers) that the dataset should not be used (e.g., see Figure \ref{fig:pwc-tiny}).

\section{Promoting Data Usability and Reproducibility}
\label{sec:datausability}

\begin{figure}
    \centering
    \includegraphics[width=0.95\textwidth]{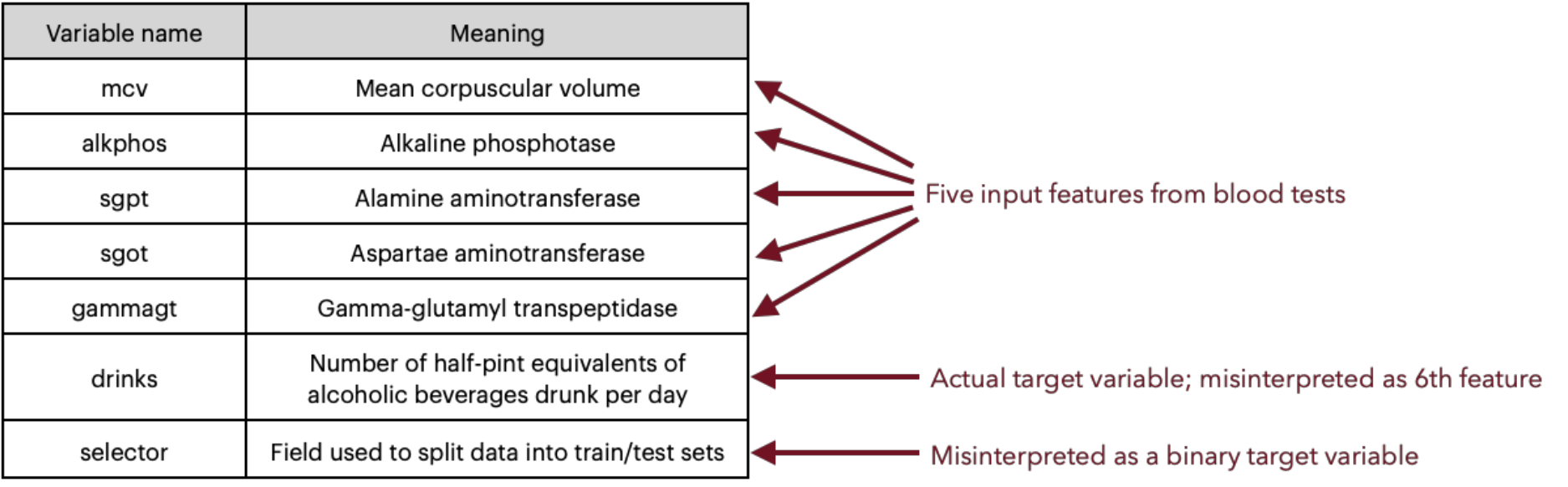}
    \caption{The BUPA Liver Disorders dataset is a popular classification benchmark from the UCI ML Repository \citep{misc_bupa_60}. Each row contains information on an individual's consumption of alcoholic drinks and their results on several blood tests targeting alcohol-related liver issues; the intended task is to predict alcohol consumption based on these test results. The last column of the dataset is an indicator, added by the dataset creators, intended to split the rows into training and test sets; however, the data documentation did not clearly explain the meaning of each column. It was subsequently found that many highly-cited papers using this dataset had mistakenly treated this last column as the class label, producing 
    ``meaningless results'' \citep{MCDERMOTT201641}.}
\label{fig:liver-disorders}
\end{figure}

Recent work has pointed to a need for improved (re)usability of data \citep{hutchinson2021towards, akhtar2024croissant} and reproducibility of benchmarked results \citep{raff2019reproducible, henderson2018deep, Friedland2024,maffia2023benchmarking,gundersen2022sources} in ML. When a dataset lacks clear metadata, it can lead to critical misunderstandings in the reuse phase of its lifecycle (as in Figure \ref{fig:liver-disorders}). Unambiguous metadata are also a necessary foundation for benchmark reproducibility, ensuring that data are used in the same way across evaluations. Benchmark reproducibility is critical for ML research: it enables the verification of published results, provides a starting point for experimentation and follow-up work, and makes contributions easier for others to use, potentially increasing research impact \cite{dehghani2021benchmark, li2023trustworthy, Tatman2018APT}.  Benchmark repositories can support usability and reproducibility with the metadata they require and provide to users \citep{hulsebos2024longer, giner2023domain}, particularly compositional and task-specific metadata (in addition to responsible points of contact and dataset versions---see Sections \ref{sec:connectionmetadata} and \ref{sec:revisiondeprecation} \citep{raff2019reproducible}). 

If repositories include benchmarked results alongside datasets (discussed further in Section \ref{sec:leaderboardculture}), they can further support reproducibility by collecting and providing metadata about the benchmarked results themselves \citep{raff2019reproducible}.

\subsection{Compositional and Task Metadata}
\label{sec:compositionaltask}
Datasets are more usable for ML researchers when accompanied by metadata describing the dataset composition and relevant tasks \citep{lin2020trust, nih2022selecting, wilkinson2016fair, oais, whostp2022desirable}. These metadata help expedite onboarding for a new dataset \citep{heger2022understanding, nair2023report}, prevent misunderstandings, and promote data reproducibility \citep{elfer2024reproducible, paullada2021data, pawlik2019link,welty2019metrology,  shavit2017stepping}.

\textit{Compositional metadata} describe the makeup of a dataset, e.g., for tabular datasets, what each instance represents, descriptions of each feature, the total number of rows and columns, the dataset label or target, the presence of missing data, and recommended data splits \citep{gebru2021datasheets, stoyanovich2019nutritional}. When this information is clear, datasets are more interoperable, meaning they can be more easily processed and incorporated into different workflows \citep{wilkinson2016fair, nair2023report}. For example, a researcher might want to evaluate an existing model on a new dataset; if this dataset has detailed compositional metadata, the researcher can quickly and easily determine which columns they need to use and what preprocessing is required. If this metadata follows a standardized schema (e.g., Croissant \citep{akhtar2024croissant}), model evaluation may even be done automatically or semi-automatically.

\textit{Task metadata} include the intended or appropriate ML tasks for a dataset (e.g., image classification or time-series prediction) and specialized metadata relevant to those tasks (e.g., for human \citep{andrews2024ethical} or medical \citep{elfer2024reproducible} image, NLP \citep{bender2018data}, or ecological \citep{shavit2017stepping} data). As an example, HuggingFace Datasets \citep{lhoest2021datasets}, which specializes in NLP data, collects metadata on language and multilinguality, text creation, and fine-grained NLP tasks (e.g., sentiment classification, multiple-choice question answering, or word sense disambiguation). Such specialized metadata make it easy for ML practitioners to find and use data that fit a specific application.

Thus, to improve data usability and benchmark reproducibility, repositories can require that data donors provide high-quality compositional and task metadata, including specialized task metadata, where appropriate
\cite{lin2020trust}. Repositories can also help streamline metadata creation processes, which donors can find overwhelming or time-consuming \citep{Xu2022cn,heger2022understanding}, e.g., with adaptive metadata collection, auto-filling or updating basic fields \citep{heger2022understanding}, or supplementary training and tools \cite{acdeb2022report}. We note that to some extent, the responsibility to provide accurate and complete metadata ultimately falls on the data donor. However, repositories can reduce the risk of incorrect, incomplete, or manipulated documentation by enforcing metadata schemata, using quality review processes, and providing user-friendly metadata creation tools---taking some of this burden from the donors themselves.

\subsection{Benchmark Metadata}
In ML benchmarking, implementation details such as software dependencies, random seeds, and hyperparameter values can have a significant impact on results \citep{dehghani2021benchmark, henderson2018deep, bouthillier2021accounting}, as can the details of metric computation, including data splits, metric definitions, and aggregation of results \citep{bischlopenml, dehghani2021benchmark}. Thus, clearly documenting this \textit{benchmark metadata} is critical for reproducibility \citep{raff2019reproducible,pineau2021improving}. Beyond validating results, the ability to replicate an analysis facilitates ``hands-on'' experimentation with benchmarked models on a given dataset, enabling researchers to test potential modifications, perform additional evaluations, or debug other models \citep{li2023trustworthy,kang2023papers}. 
To this end, if a repository displays a particular benchmarked result for a dataset, they should ensure that specific details on all settings and hyperparameter values used to obtain that result are available \citep{raff2019reproducible}. The software environments, dependencies, code, and data files necessary to re-run analyses should also be documented and accessible \citep{Sawchuk2021qy, heger2022understanding, gundersen2021fundamental,li2023trustworthy, laurinavichyute2022share, wang2021kaggle}. By ensuring that detailed metadata on a dataset's content, composition, task, and benchmarked results are available, repositories can provide ML practitioners with a holistic understanding of the benchmark \citep{leipzig2021role, heger2022understanding}.

\section{Encouraging Holistic Evaluation}
\label{sec:leaderboardculture}
 
It has become common practice to tabulate benchmarked metrics for different ML methods with a dataset leaderboard; several benchmark repositories offer leaderboard features, including Kaggle \citep{banachewicz2022kaggle}
and Papers with Code.\footnote{e.g., \href{https://paperswithcode.com/sota/image-classification-on-cifar-10}{https://paperswithcode.com/sota/image-classification-on-cifar-10}} 
Leaderboarding has become predominant in ML evaluation, and state-of-the-art performance is a key factor in peer review processes \citep{dehghani2021benchmark, teney2020value}.
However, this leaderboard culture has been criticized for a ``near singular'' focus on the incremental improvement of a narrow set of metrics (e.g., classification accuracy) \citep{paullada2021data, thomas2022reliance}.  
Such fixation on a specific metric is unlikely to yield broadly applicable results and can stifle the growth of new, diverse ideas \citep{paullada2021data, dehghani2021benchmark, raji2021ai, teney2020value}. Further, as measures of uncertainty are seldom incorporated, seemingly record-breaking performance improvements are not always statistically significant \citep{teney2020value, everingham2015pascal,fernandez2014hundreds} (e.g., a review of the MS MARCO leaderboard found no significant difference in performance between the top three models \citep{lin2021significant}).
In this paper, we refrain from taking a stance on whether benchmark repositories should include leaderboards. However, as several repositories currently act as centralized purveyors of leaderboards, we briefly discuss how they can promote more comprehensive evaluation, helping address problems with benchmarking that manifest later in the dataset lifecycle.

\subsection{Analysis Beyond Single Metrics}
Recent work on best practices for ML benchmarking recommends evaluating performance more holistically \citep{raji2021ai, paullada2021data, burnell2023rethink, ribeiro2020beyond}. This could include a variety of metrics, capturing model size and complexity, energy consumption, inference latency, and the amount of data used \citep{Hoffmann2019yi,ethayarajh2020utility, Thiyagalingam2022qt, dehghani2021benchmark, thomas2022reliance, lipton2019troubling}. 
Additional in-depth assessment---such as error analysis or disaggregated evaluations---can provide a more nuanced portrait of model behavior, capturing bias, fairness, or robustness \citep{raji2021ai, wu2019errudite, paullada2021data, dotan2020value, bowman2021fix}.
Thus, to enhance their leaderboards, benchmark repositories can include this sort of comprehensive information on model performance. 
In addition to incentivizing progress in a number of dimensions, this approach reflects that the ideal model is context-dependent, enabling practitioners to choose a model based on the criteria most relevant to their use case \citep{paullada2021data, dehghani2021benchmark, dotan2020value, ethayarajh2020utility}.

\subsection{Metric Uncertainty}
Metrics shown without any measure of uncertainty can prompt fallacious conclusions, e.g., that one model performs definitively better than another. Instead, including uncertainty makes these benchmarked results more informative 
\cite{totsch2021, Thiyagalingam2022qt}, and a growing body of methodologies has developed for estimating uncertainty, computing confidence intervals, and performing statistical significance testing in the context of model comparison \cite{graham2014randomized,dror2017replicability, bouthillier2021accounting, lin2021significant,dehghani2021benchmark,wein2023follow}. In light of this, several guidelines for ML evaluation call for the inclusion of variance, uncertainty, and statistical significance in model analysis \citep{hand2006classifier, teney2020value, Mllersen2023WhatIT, biderman2020pitfalls, bowman2021fix}. Repositories with leaderboards can support this movement by enforcing the reporting of uncertainty alongside point estimates of metrics.

\section{Diversifying Benchmark Datasets}
\label{sec:benchmarkoveruse}
Benchmarking for a particular type of ML model is often concentrated on a limited set of datasets and tasks \citep{dotan2020value, raji2021ai, koch2021reduced}. 
This lack of diversity can encourage overfitting to a specific benchmark dataset (e.g., via random seed or hyperparameter fishing) \citep{pmlrrecht19a, Mllersen2023WhatIT}. Over-adaptation also happens at a macro level over time, as new models leverage tricks and strategies from earlier work \citep{hand2006classifier, dehghani2021benchmark}. 
Moreover, these benchmarks are often not directly relevant to the real-world behavior they are evaluating---for example, the GLUE benchmark \citep{wang2019glue} has been commonly used to evaluate natural language understanding, but it is mainly comprised of sequence matching tasks \citep{dehghani2021benchmark}. As a result, there is often a disconnect between benchmarked performance and real-world model behavior \citep{matters2012wagstaff, mccoy2019right, hand2006classifier}.
Thus, the overrepresentation of specific tasks, data types, and test datasets can ultimately bias long-term research directions \citep{dehghani2021benchmark, raji2021ai, dotan2020value} and limit the generalizability of model evaluations \citep{denton2020bringing, biderman2020pitfalls, bowman2021fix, schlangen2021targeting}. To fight these patterns of overfitting and overuse, repositories can support the discovery and use of diverse, relevant, and continuously evolving datasets.

\subsection{Living Datasets}
To hinder the overfitting of models to a specific test set, leaderboards can evaluate submitted models on a private, hitherto unused test set \cite{dwork2015preserving,blum2015ladder,roelofs2019meta} (e.g., as done by Kaggle) or on out-of-distribution data \citep{koh2021wilds, teney2020value}. Extending this principle, leaderboards can also support ``living'' or evolving datasets, to which dataset creators continuously add new examples or tasks and remove outdated or erroneous examples. While this means that the benchmarked performances of two models evaluated at two different points in time may not be directly comparable (and data versioning, as discussed in Section \ref{sec:revisiondeprecation}, is critical), robust models will generally outperform those using a specific trick or artifact as evaluations are repeated over time \citep{dehghani2021benchmark}. These living datasets also track the real-world evolution of data---for instance, in the context of autonomous driving, new types of vehicles appear on the roads \citep{liangadvances2022}---which static benchmark datasets fail to capture \citep{sasha2021framework}. By evaluating models on living datasets, repositories can help shift focus away from a specific static set of examples, de-incentivizing overfitting and helping bridge the gap between benchmarked and real-world performance.

\subsection{Dataset Discoverability}
When choosing benchmark datasets for evaluating an algorithm, it has become the default to select the same datasets already used in the literature  \citep{denton2020bringing, raji2021face}. Often, however, there also exists a plethora of other high-quality datasets that could have been used but did not win the ``benchmark lottery'' \cite{dehghani2021benchmark} and were left undiscovered. 
To support \textit{dataset discoverability}, existing standards emphasize the importance of standardized, rich metadata \citep{lin2020trust, nih2022selecting, whostp2022desirable}, which enable searching for datasets via keywords, filtering, and controlled vocabularies \cite{datacite}.

For benchmark repositories, this search is often task-driven: ML practitioners need to find datasets for which a certain type of model is applicable, based on compositional properties and relevant tasks (see Section \ref{sec:compositionaltask}). 
Thus, to improve benchmark dataset discovery, repositories can support search based on compositional and task metadata \citep{heger2022understanding, hulsebos2024longer, desai2024archival}---for example, the UCI ML Repository's search functionality includes a filter for classification, regression, clustering, or other datasets. 
Overall, by promoting the discovery and use of a more diverse set of evaluation datasets, repositories can build a barrier to the over-representation of specific benchmark tasks or datasets and encourage more generalizable model evaluation.

\section{Discussion and Conclusion}
\label{sec:discussion}

\subsection{Key Takeaways}
A common thread throughout the criticisms of ML data and benchmarking practices we discuss in this paper is a need for the intervention of a third party---separate from dataset creators and users---in addressing these issues \citep{sambasivan2021everyone, jo_lessons_2020, desai2024archival, hutchinson2021towards, paullada2021data, dehghani2021benchmark, scheuerman2021datasets, biderman2020pitfalls, heinzerling2020nlp, andrews2024ethical, raji2021ai, sasha2021framework, heger2022understanding, lin2020trust}. While improving the state of ML evaluation will be a community effort, involving the efforts of conferences and journals, policymakers, nonprofit organizations, and individual practitioners \citep{jo_lessons_2020}, in this paper we posit that benchmark repositories can play a major role in this effort, instigating far-reaching changes to the culture surrounding datasets and benchmarking in ML. We summarize our key takeaways as follows.
\begin{itemize}
    \item Repositories can highlight the status of datasets as valuable scholarly contributions.
    \item Repositories are well-positioned in the data pipeline to address issues with dataset content.
    \item Repositories can facilitate data reuse and benchmark reproducibility by ensuring salient metadata is provided for datasets (and, if applicable, benchmark evaluations).
    \item Repositories with leaderboard features can enforce best practices for model evaluation.
    \item Repositories can provide a platform for discovering new, relevant, high-quality datasets, counteracting the overuse of a small set of standard benchmark datasets.
\end{itemize}

\subsection{Limitations}
The long-term feasibility and impact of our suggestions are predicated upon larger shifts in community norms and attitudes about data-centric work, which will rely upon proper incentivization, an open challenge in the ML community \citep{gero2023incentive, jo_lessons_2020, hutchinson2021towards}. 
We discuss here potential incentives for both individual researchers and repositories---although incentives for other actors (e.g., universities, companies, publishers) are also worth exploring.

A key incentive for researchers to become involved in repository efforts is funding; this is becoming more available as agencies such as the U.S. National Institutes of Health (NIH) and National Science Foundation (NSF) pay increasing attention to the data-sharing ecosystem.\footnote{\href{https://grants.nih.gov/grants/guide/notice-files/NOT-OD-21-013.html}{https://grants.nih.gov/grants/guide/notice-files/NOT-OD-21-013.html}} For example, the NSF’s program for Community Infrastructure for Research in Computer and Information Science and Engineering explicitly calls out funding support for data repositories\footnote{\href{https://new.nsf.gov/funding/opportunities/circ-community-infrastructure-research-computer-information/nsf23-589/solicitation}{https://new.nsf.gov/funding/opportunities/circ-community-infrastructure-research-computer-information/nsf23-589/solicitation}}, and the U.S. National Artificial Intelligence Research Resource Task Force identifies repositories as important to their goal of ``Strengthening and Democratizing the U.S. Artificial Intelligence Innovation Ecosystem'' \cite{nair2023report}. 

To incentivize dataset reviewers, repositories could follow the model of ``volunteer journals'' such as the \textit{Journal of Machine Learning Research}These public journals demonstrate how high-quality shared resources can be developed through dedicated volunteer efforts, offering inspiration for a parallel system of oversight, reviewing, and maintenance for repositories. For example, similar to the role of Action or Associate Editors (AEs) in these journals, repositories could have a set of curators who are responsible for identifying relevant experts to review a dataset. By framing data curation and review as an academic service in the same vein as more traditional editorial roles, repositories could help incentivize participation in the review process.

Further work is also needed to determine how to incentivize repositories themselves to enforce best practices (e.g., requiring data donors to select a license or provide task metadata). One potential avenue is to establish standards for benchmark repositories, building upon standards for data repositories in general, such as CoreTrustSeal\footnote{\href{https://www.coretrustseal.org/}{https://www.coretrustseal.org/}}. Establishing standards or repository certification processes will be most effective if the ML community cultivates an expectation that such requirements are met (e.g., as in archival settings \cite{jo_lessons_2020}).

Though we may draw useful inspiration from these ideas, it remains unclear what incentivization strategies will work best to spur large-scale buy-in from repositories, dataset creators, and dataset users in implementing and maintaining best practices.

\subsection{Looking Ahead}

Going forward, we hope that a growing appreciation of data work will permeate the ML community, serving as a catalyst for investment into data infrastructure in ML and broader researcher involvement in data repositories. 
In light of this, we believe the ideas in this paper lay a foundation for further discussion and research about how benchmark repositories can be utilized, and improved, for better benchmarking in ML.

\section*{Acknowledgements}
This research was supported in part by the National Science Foundation under award CCRI-1925741 and in part by the Hasso Plattner Institute (HPI) Research Center in Machine Learning and Data Science at the University of California, Irvine.

\bibliographystyle{unsrtnat}
\bibliography{refs}

\begin{thebibliography}{174}
\providecommand{\natexlab}[1]{#1}
\providecommand{\url}[1]{\texttt{#1}}
\expandafter\ifx\csname urlstyle\endcsname\relax
  \providecommand{\doi}[1]{doi: #1}\else
  \providecommand{\doi}{doi: \begingroup \urlstyle{rm}\Url}\fi

\bibitem[Thiyagalingam et~al.(2022)Thiyagalingam, Shankar, Fox, and Hey]{Thiyagalingam2022qt}
Jeyan Thiyagalingam, Mallikarjun Shankar, Geoffrey Fox, and Tony Hey.
\newblock Scientific machine learning benchmarks.
\newblock \emph{Nature Reviews Physics}, 4\penalty0 (6):\penalty0 413--420, April 2022.

\bibitem[Bent{\'e}jac et~al.(2021)Bent{\'e}jac, Cs{\"o}rg{\H{o}}, and Mart{\'\i}nez-Mu{\~n}oz]{bentejac2021comparative}
Candice Bent{\'e}jac, Anna Cs{\"o}rg{\H{o}}, and Gonzalo Mart{\'\i}nez-Mu{\~n}oz.
\newblock A comparative analysis of gradient boosting algorithms.
\newblock \emph{Artificial Intelligence Review}, 54:\penalty0 1937--1967, 2021.

\bibitem[Hoffmann et~al.(2019)Hoffmann, Bertram, Mikut, Reischl, and Nelles]{Hoffmann2019yi}
Frank Hoffmann, Torsten Bertram, Ralf Mikut, Markus Reischl, and Oliver Nelles.
\newblock Benchmarking in classification and regression.
\newblock \emph{WIREs Data Mining and Knowledge Discovery}, 9\penalty0 (5):\penalty0 e1318, September 2019.

\bibitem[Caruana and Niculescu-Mizil(2006)]{caruana2006empirical}
Rich Caruana and Alexandru Niculescu-Mizil.
\newblock An empirical comparison of supervised learning algorithms.
\newblock In \emph{Proceedings of the 2006 International Conference on Machine Learning}, pages 161--168, 2006.

\bibitem[Dehghani et~al.(2021)Dehghani, Tay, Gritsenko, Zhao, Houlsby, Diaz, Metzler, and Vinyals]{dehghani2021benchmark}
Mostafa Dehghani, Yi~Tay, Alexey~A Gritsenko, Zhe Zhao, Neil Houlsby, Fernando Diaz, Donald Metzler, and Oriol Vinyals.
\newblock The benchmark lottery.
\newblock \emph{arXiv preprint arXiv:2107.07002}, 2021.

\bibitem[Raji et~al.(2021)Raji, Bender, Paullada, Denton, and Hanna]{raji2021ai}
Inioluwa~Deborah Raji, Emily~M Bender, Amandalynne Paullada, Emily Denton, and Alex Hanna.
\newblock A{I} and the everything in the whole wide world benchmark.
\newblock In \emph{35th Conference on Neural Information Processing Systems (NeurIPS 2021) Track on Datasets and Benchmarks}, 2021.
\newblock URL \url{https://openreview.net/pdf?id=j6NxpQbREA1}.

\bibitem[Paullada et~al.(2021)Paullada, Raji, Bender, Denton, and Hanna]{paullada2021data}
Amandalynne Paullada, Inioluwa~Deborah Raji, Emily~M Bender, Emily Denton, and Alex Hanna.
\newblock Data and its (dis) contents: A survey of dataset development and use in machine learning research.
\newblock \emph{Patterns}, 2\penalty0 (11), 2021.

\bibitem[Dotan and Milli(2020)]{dotan2020value}
Ravit Dotan and Smitha Milli.
\newblock Value-laden disciplinary shifts in machine learning.
\newblock In \emph{Proceedings of the 2020 Conference on Fairness, Accountability, and Transparency}, FAT* '20, page 294, New York, NY, USA, 2020. Association for Computing Machinery.
\newblock URL \url{https://doi.org/10.1145/3351095.3373157}.

\bibitem[Denton et~al.(2020)Denton, Hanna, Amironesei, Smart, Nicole, and Scheuerman]{denton2020bringing}
Emily Denton, Alex Hanna, Razvan Amironesei, Andrew Smart, Hilary Nicole, and Morgan~Klaus Scheuerman.
\newblock Bringing the people back in: Contesting benchmark machine learning datasets.
\newblock \emph{arXiv preprint arXiv:2007.07399}, 2020.

\bibitem[Wagstaff(2012)]{matters2012wagstaff}
Kiri~L. Wagstaff.
\newblock Machine learning that matters.
\newblock In \emph{Proceedings of the 2012 International Conference on Machine Learning}, page 1851–1856, Madison, WI, USA, 2012. Omnipress.

\bibitem[Hicks and Irizarry(2018)]{hicks2018teaching}
Stephanie~C. Hicks and Rafael~A. Irizarry.
\newblock A guide to teaching data science.
\newblock \emph{The American Statistician}, 72\penalty0 (4):\penalty0 382--391, 2018.
\newblock URL \url{https://doi.org/10.1080/00031305.2017.1356747}.

\bibitem[Serrano et~al.(2017)Serrano, Molina, Manrique, and Baumela]{serrano2017experiential}
Emilio Serrano, Martin Molina, Daniel Manrique, and Luis Baumela.
\newblock Experiential learning in data science: {F}rom the dataset repository to the platform of experiences.
\newblock \emph{Intelligent Environments 2017}, 22:\penalty0 122--130, 2017.

\bibitem[Langley(2011)]{langley2011changing}
Pat Langley.
\newblock The changing science of machine learning.
\newblock \emph{Machine Learning}, 82\penalty0 (3):\penalty0 275--279, 2011.

\bibitem[Vanschoren et~al.(2014)Vanschoren, Van~Rijn, Bischl, and Torgo]{vanschoren2014openml}
Joaquin Vanschoren, Jan~N Van~Rijn, Bernd Bischl, and Luis Torgo.
\newblock Open{ML}: {N}etworked science in machine learning.
\newblock \emph{ACM SIGKDD Explorations Newsletter}, 15\penalty0 (2):\penalty0 49--60, 2014.

\bibitem[Bischl et~al.(2021)Bischl, Casalicchio, Feurer, Gijsbers, Hutter, Lang, Mantovani, van Rijn, and Vanschoren]{bischlopenml}
Bernd Bischl, Giuseppe Casalicchio, Matthias Feurer, Pieter Gijsbers, Frank Hutter, Michel Lang, Rafael~G Mantovani, Jan~N van Rijn, and Joaquin Vanschoren.
\newblock Open{ML} benchmarking suites.
\newblock In \emph{35th Conference on Neural Information Processing Systems (NeurIPS 2021): Track on Datasets and Benchmarks}, 2021.

\bibitem[Brazdil et~al.(2022)Brazdil, Rijn, Soares, and Vanschoren]{brazdil2022metadata}
Pavel Brazdil, Jan N~van Rijn, Carlos Soares, and Joaquin Vanschoren.
\newblock Metadata repositories.
\newblock In \emph{Metalearning}, chapter~16, pages 297--310. Springer, 2022.

\bibitem[Lhoest et~al.(2021)Lhoest, Villanova~del Moral, Jernite, Thakur, von Platen, Patil, Chaumond, Drame, Plu, Tunstall, Davison, {\v{S}}a{\v{s}}ko, Chhablani, Malik, Brandeis, Le~Scao, Sanh, Xu, Patry, McMillan-Major, Schmid, Gugger, Delangue, Matussi{\`e}re, Debut, Bekman, Cistac, Goehringer, Mustar, Lagunas, Rush, and Wolf]{lhoest2021datasets}
Quentin Lhoest, Albert Villanova~del Moral, Yacine Jernite, Abhishek Thakur, Patrick von Platen, Suraj Patil, Julien Chaumond, Mariama Drame, Julien Plu, Lewis Tunstall, Joe Davison, Mario {\v{S}}a{\v{s}}ko, Gunjan Chhablani, Bhavitvya Malik, Simon Brandeis, Teven Le~Scao, Victor Sanh, Canwen Xu, Nicolas Patry, Angelina McMillan-Major, Philipp Schmid, Sylvain Gugger, Cl{\'e}ment Delangue, Th{\'e}o Matussi{\`e}re, Lysandre Debut, Stas Bekman, Pierric Cistac, Thibault Goehringer, Victor Mustar, Fran{\c{c}}ois Lagunas, Alexander Rush, and Thomas Wolf.
\newblock Datasets: A community library for natural language processing.
\newblock In \emph{Proceedings of the 2021 Conference on Empirical Methods in Natural Language Processing: System Demonstrations}, pages 175--184, Online and Punta Cana, Dominican Republic, November 2021. Association for Computational Linguistics.
\newblock URL \url{https://aclanthology.org/2021.emnlp-demo.21}.

\bibitem[Banachewicz and Massaron(2022)]{banachewicz2022kaggle}
Konrad Banachewicz and Luca Massaron.
\newblock \emph{The Kaggle Book: Data analysis and machine learning for competitive data science}.
\newblock Packt Publishing Ltd, 2022.

\bibitem[Dau et~al.(2019)Dau, Bagnall, Kamgar, Yeh, Zhu, Gharghabi, Ratanamahatana, and Keogh]{dau2019ucr}
Hoang~Anh Dau, Anthony Bagnall, Kaveh Kamgar, Chin-Chia~Michael Yeh, Yan Zhu, Shaghayegh Gharghabi, Chotirat~Ann Ratanamahatana, and Eamonn Keogh.
\newblock The {U}{C}{R} time series archive.
\newblock \emph{IEEE/CAA Journal of Automatica Sinica}, 6\penalty0 (6):\penalty0 1293--1305, 2019.

\bibitem[Olson et~al.(2017)Olson, La~Cava, Orzechowski, Urbanowicz, and Moore]{olson2017pmlb}
Randal~S Olson, William La~Cava, Patryk Orzechowski, Ryan~J Urbanowicz, and Jason~H Moore.
\newblock {P}{M}{L}{B}: a large benchmark suite for machine learning evaluation and comparison.
\newblock \emph{BioData mining}, 10:\penalty0 1--13, 2017.

\bibitem[Ashmore et~al.(2021)Ashmore, Calinescu, and Paterson]{ashmore2021assuring}
Rob Ashmore, Radu Calinescu, and Colin Paterson.
\newblock Assuring the machine learning lifecycle: Desiderata, methods, and challenges.
\newblock \emph{ACM Computing Surveys}, 54\penalty0 (5), May 2021.
\newblock URL \url{https://doi.org/10.1145/3453444}.

\bibitem[Park and Cordell(2023)]{park2023ripple}
Jaihyun Park and Ryan Cordell.
\newblock The ripple effect of dataset reuse: Contextualising the data lifecycle for machine learning data sets and social impact.
\newblock \emph{Journal of Information Science}, 2023.
\newblock URL \url{https://doi.org/10.1177/01655515231212977}.

\bibitem[Koch et~al.(2021)Koch, Denton, Hanna, and Foster]{koch2021reduced}
Bernard Koch, Emily Denton, Alex Hanna, and Jacob~Gates Foster.
\newblock Reduced, reused and recycled: The life of a dataset in machine learning research.
\newblock In \emph{35th Conference on Neural Information Processing Systems (NeurIPS 2021): Track on Datasets and Benchmarks}, 2021.
\newblock URL \url{https://openreview.net/forum?id=zNQBIBKJRkd}.

\bibitem[Gero et~al.(2023)Gero, Das, Dognin, Padhi, Sattigeri, and Varshney]{gero2023incentive}
Katy~Ilonka Gero, Payel Das, Pierre Dognin, Inkit Padhi, Prasanna Sattigeri, and Kush~R Varshney.
\newblock The incentive gap in data work in the era of large models.
\newblock \emph{Nature Machine Intelligence}, 5\penalty0 (6):\penalty0 565--567, 2023.

\bibitem[Birhane et~al.(2022)Birhane, Kalluri, Card, Agnew, Dotan, and Bao]{birhane2022values}
Abeba Birhane, Pratyusha Kalluri, Dallas Card, William Agnew, Ravit Dotan, and Michelle Bao.
\newblock The values encoded in machine learning research.
\newblock In \emph{Proceedings of the 2022 ACM Conference on Fairness, Accountability, and Transparency (FAccT)}, pages 173--184, 2022.

\bibitem[Sambasivan et~al.(2021)Sambasivan, Kapania, Highfill, Akrong, Paritosh, and Aroyo]{sambasivan2021everyone}
Nithya Sambasivan, Shivani Kapania, Hannah Highfill, Diana Akrong, Praveen Paritosh, and Lora~M Aroyo.
\newblock “{E}veryone wants to do the model work, not the data work”: {D}ata cascades in high-stakes {A}{I}.
\newblock In \emph{Proceedings of the 2021 CHI Conference on Human Factors in Computing Systems}, pages 1--15, 2021.

\bibitem[Jo and Gebru(2020)]{jo_lessons_2020}
Eun~Seo Jo and Timnit Gebru.
\newblock Lessons from archives: {S}trategies for collecting sociocultural data in machine learning.
\newblock In \emph{Proceedings of the 2020 {Conference} on {Fairness}, {Accountability}, and {Transparency}}, {FAT}* '20, pages 306--316, New York, NY, USA, January 2020. Association for Computing Machinery.
\newblock URL \url{https://dl.acm.org/doi/10.1145/3351095.3372829}.

\bibitem[Hutchinson et~al.(2021)Hutchinson, Smart, Hanna, Denton, Greer, Kjartansson, Barnes, and Mitchell]{hutchinson2021towards}
Ben Hutchinson, Andrew Smart, Alex Hanna, Emily Denton, Christina Greer, Oddur Kjartansson, Parker Barnes, and Margaret Mitchell.
\newblock Towards accountability for machine learning datasets: Practices from software engineering and infrastructure.
\newblock In \emph{Proceedings of the 2021 ACM Conference on Fairness, Accountability, and Transparency (FAccT)}, page 560–575, New York, NY, USA, 2021. Association for Computing Machinery.
\newblock URL \url{https://doi.org/10.1145/3442188.3445918}.

\bibitem[Longpre et~al.(2024)Longpre, Yauney, Reif, Lee, Roberts, Zoph, Zhou, Wei, Robinson, Mimno, and Ippolito]{longpre2023pretrainer}
Shayne Longpre, Gregory Yauney, Emily Reif, Katherine Lee, Adam Roberts, Barret Zoph, Denny Zhou, Jason Wei, Kevin Robinson, David Mimno, and Daphne Ippolito.
\newblock A pretrainer{'}s guide to training data: Measuring the effects of data age, domain coverage, quality, {\&} toxicity.
\newblock In Kevin Duh, Helena Gomez, and Steven Bethard, editors, \emph{Proceedings of the 2024 Conference of the North American Chapter of the Association for Computational Linguistics: Human Language Technologies (Volume 1: Long Papers)}, pages 3245--3276, Mexico City, Mexico, June 2024. Association for Computational Linguistics.
\newblock \doi{10.18653/v1/2024.naacl-long.179}.
\newblock URL \url{https://aclanthology.org/2024.naacl-long.179}.

\bibitem[Bhardwaj et~al.(2024)Bhardwaj, Gujral, Wu, Zogheib, Maharaj, and Becker]{bhardwaj2024machine}
Eshta Bhardwaj, Harshit Gujral, Siyi Wu, Ciara Zogheib, Tegan Maharaj, and Christoph Becker.
\newblock Machine learning data practices through a data curation lens: {A}n evaluation framework.
\newblock In \emph{Proceedings of the 2024 ACM Conference on Fairness, Accountability, and Transparency (FAccT)}, FAccT '24, New York, NY, USA, 2024. Association for Computing Machinery.

\bibitem[Heinzerling(2020)]{heinzerling2020nlp}
Benjamin Heinzerling.
\newblock {N}{L}{P}'s clever hans moment has arrived.
\newblock \emph{Journal of Cognitive Science}, 21\penalty0 (1), 2020.

\bibitem[Scheuerman et~al.(2021)Scheuerman, Hanna, and Denton]{scheuerman2021datasets}
Morgan~Klaus Scheuerman, Alex Hanna, and Emily Denton.
\newblock Do datasets have politics? {D}isciplinary values in computer vision dataset development.
\newblock In \emph{Proceedings of the ACM on Human-Computer Interaction}, volume~5, pages 1--37. ACM Press, New York, NY, USA, 2021.

\bibitem[Desai et~al.(2024)Desai, Pasquetto, Jacobs, and Card]{desai2024archival}
Meera~A. Desai, Irene~V. Pasquetto, Abigail~Z. Jacobs, and Dallas Card.
\newblock An archival perspective on pretraining data.
\newblock \emph{Patterns}, 5\penalty0 (4):\penalty0 100966, 2024.
\newblock URL \url{https://doi.org/10.1016/j.patter.2024.100966}.

\bibitem[Oala et~al.(2024)Oala, Maskey, Bat-Leah, Parrish, G{\"u}rel, Kuo, Liu, Dror, Brajovic, Yao, Bartolo, Rojas, Hileman, Aliment, Mahoney, Risdal, Lease, Samek, Dutta, Northcutt, Coleman, Hancock, Koch, Tadesse, Karla{\v{s}}, Alaa, Dieng, Noy, Reddi, Zou, Paritosh, van~der Schaar, Bollacker, Aroyo, Zhang, Vanschoren, Guyon, and Mattson]{oala2024dmlr}
Luis Oala, Manil Maskey, Lilith Bat-Leah, Alicia Parrish, Nezihe~Merve G{\"u}rel, Tzu-Sheng Kuo, Yang Liu, Rotem Dror, Danilo Brajovic, Xiaozhe Yao, Max Bartolo, William A~Gaviria Rojas, Ryan Hileman, Rainier Aliment, Michael~W. Mahoney, Meg Risdal, Matthew Lease, Wojciech Samek, Debojyoti Dutta, Curtis~G Northcutt, Cody Coleman, Braden Hancock, Bernard Koch, Girmaw~Abebe Tadesse, Bojan Karla{\v{s}}, Ahmed Alaa, Adji~Bousso Dieng, Natasha Noy, Vijay~Janapa Reddi, James Zou, Praveen Paritosh, Mihaela van~der Schaar, Kurt Bollacker, Lora Aroyo, Ce~Zhang, Joaquin Vanschoren, Isabelle Guyon, and Peter Mattson.
\newblock {D}{M}{L}{R}: Data-centric machine learning research - past, present and future.
\newblock \emph{Journal of Data-centric Machine Learning Research}, 2024.
\newblock URL \url{https://openreview.net/forum?id=2kpu78QdeE}.

\bibitem[Wilkinson et~al.(2016)Wilkinson, Dumontier, Aalbersberg, Appleton, Axton, Baak, Blomberg, Boiten, Bonino~da Silva~Santos, Bourne, Bouwman, Brookes, Clark, Crosas, Dillo, Dumon, Edmunds, Evelo, Finkers, and Mons]{wilkinson2016fair}
Mark Wilkinson, Michel Dumontier, IJsbrand~Jan Aalbersberg, Gaby Appleton, Myles Axton, Arie Baak, Niklas Blomberg, Jan-Willem Boiten, Luiz~Olavo Bonino~da Silva~Santos, Philip Bourne, Jildau Bouwman, Anthony Brookes, Tim Clark, Merce Crosas, Ingrid Dillo, Olivier Dumon, Scott Edmunds, Chris Evelo, Richard Finkers, and Barend Mons.
\newblock The {FAIR} guiding principles for scientific data management and stewardship.
\newblock \emph{Scientific Data}, 3, 03 2016.

\bibitem[Stodden et~al.(2016)Stodden, McNutt, Bailey, Deelman, Gil, Hanson, Heroux, Ioannidis, and Taufer]{stodden2016enhancing}
Victoria Stodden, Marcia McNutt, David~H Bailey, Ewa Deelman, Yolanda Gil, Brooks Hanson, Michael~A Heroux, John~PA Ioannidis, and Michela Taufer.
\newblock Enhancing reproducibility for computational methods.
\newblock \emph{Science}, 354\penalty0 (6317):\penalty0 1240--1241, 2016.

\bibitem[Goodman et~al.(2014)Goodman, Pepe, Blocker, Borgman, Cranmer, Crosas, Di~Stefano, Gil, Groth, Hedstrom, et~al.]{goodman2014ten}
Alyssa Goodman, Alberto Pepe, Alexander~W Blocker, Christine~L Borgman, Kyle Cranmer, Merce Crosas, Rosanne Di~Stefano, Yolanda Gil, Paul Groth, Margaret Hedstrom, et~al.
\newblock Ten simple rules for the care and feeding of scientific data.
\newblock \emph{PLoS Computational Biology}, 10\penalty0 (4):\penalty0 e1003542, 2014.

\bibitem[Liang et~al.(2022)Liang, Tadesse, Ho, Fei-Fei, Zaharia, Zhang, and Zou]{liangadvances2022}
Weixin Liang, Girmaw~Abebe Tadesse, Daniel Ho, L.~Fei-Fei, Matei Zaharia, Ce~Zhang, and James Zou.
\newblock Advances, challenges and opportunities in creating data for trustworthy {A}{I}.
\newblock \emph{Nature Machine Intelligence}, 4\penalty0 (8):\penalty0 669–677, Aug 2022.

\bibitem[{National Science and Technology Council}(2022)]{whostp2022desirable}
{National Science and Technology Council}.
\newblock Desirable characteristics of data repositories for federally funded research.
\newblock \url{https://doi.org/10.5479/10088/113528}, 2022.

\bibitem[Peng et~al.(2021)Peng, Mathur, and Narayanan]{peng2021mitigating}
Kenneth Peng, Arunesh Mathur, and Arvind Narayanan.
\newblock Mitigating dataset harms requires stewardship: {L}essons from 1000 papers.
\newblock In \emph{Proceedings of the Neural Information Processing Systems Track on Datasets and Benchmarks}, volume~1. Curran Associates, Inc., 2021.

\bibitem[Heger et~al.(2022)Heger, Marquis, Vorvoreanu, Wallach, and Wortman~Vaughan]{heger2022understanding}
Amy~K. Heger, Liz~B. Marquis, Mihaela Vorvoreanu, Hanna Wallach, and Jennifer Wortman~Vaughan.
\newblock Understanding machine learning practitioners' data documentation perceptions, needs, challenges, and desiderata.
\newblock \emph{Proc. ACM Hum.-Comput. Interact.}, 6\penalty0 (CSCW2), Nov 2022.
\newblock URL \url{https://doi.org/10.1145/3555760}.

\bibitem[on~Data Citation~Standards and Practices(2013)]{task2013out}
Task~Group on~Data Citation~Standards and CODATA-ICSTI Practices.
\newblock Out of cite, out of mind: The current state of practice, policy, and technology for the citation of data.
\newblock \emph{Data Science Journal}, 12\penalty0 (0):\penalty0 CIDCR1--CIDCR75, 2013.

\bibitem[{Data Citation Synthesis Group}(2014)]{jddcp}
{Data Citation Synthesis Group}.
\newblock Joint declaration of data citation principles, 2014.
\newblock URL \url{https://doi.org/10.25490/a97f-egyk}.

\bibitem[Altman et~al.(2015)Altman, Borgman, Crosas, and Matone]{altman2015introduction}
Micah Altman, Christine Borgman, Merc{\`e} Crosas, and Maryann Matone.
\newblock An introduction to the joint principles for data citation.
\newblock \emph{Bulletin of the Association for Information Science and Technology}, 41\penalty0 (3):\penalty0 43--45, 2015.

\bibitem[Silvello(2018)]{silvello2018theory}
Gianmaria Silvello.
\newblock Theory and practice of data citation.
\newblock \emph{Journal of the Association for Information Science and Technology}, 69\penalty0 (1):\penalty0 6--20, 2018.

\bibitem[Groth et~al.(2020)Groth, Cousijn, Clark, and Goble]{groth2020fair}
Paul Groth, Helena Cousijn, Tim Clark, and Carole Goble.
\newblock Fair data reuse--the path through data citation.
\newblock \emph{Data Intelligence}, 2\penalty0 (1-2):\penalty0 78--86, 2020.

\bibitem[Fenner et~al.(2019)Fenner, Crosas, Grethe, Kennedy, Hermjakob, Rocca-Serra, Durand, Berjon, Karcher, Martone, et~al.]{fenner2019data}
Martin Fenner, Merc{\`e} Crosas, Jeffrey~S Grethe, David Kennedy, Henning Hermjakob, Phillippe Rocca-Serra, Gustavo Durand, Robin Berjon, Sebastian Karcher, Maryann Martone, et~al.
\newblock A data citation roadmap for scholarly data repositories.
\newblock \emph{Scientific Data}, 6\penalty0 (1):\penalty0 28, 2019.

\bibitem[Starr et~al.(2015)Starr, Castro, Crosas, Dumontier, Downs, Duerr, Haak, Haendel, Herman, Hodson, et~al.]{starr2015achieving}
Joan Starr, Eleni Castro, Merc{\`e} Crosas, Michel Dumontier, Robert~R Downs, Ruth Duerr, Laurel~L Haak, Melissa Haendel, Ivan Herman, Simon Hodson, et~al.
\newblock Achieving human and machine accessibility of cited data in scholarly publications.
\newblock \emph{PeerJ Computer Science}, 1:\penalty0 e1, 2015.

\bibitem[Cousijn et~al.(2018)Cousijn, Kenall, Ganley, Harrison, Kernohan, Lemberger, Murphy, Polischuk, Taylor, Martone, et~al.]{cousijn2018data}
Helena Cousijn, Amye Kenall, Emma Ganley, Melissa Harrison, David Kernohan, Thomas Lemberger, Fiona Murphy, Patrick Polischuk, Simone Taylor, Maryann Martone, et~al.
\newblock A data citation roadmap for scientific publishers.
\newblock \emph{Scientific Data}, 5\penalty0 (1):\penalty0 1--11, 2018.

\bibitem[Lowenberg et~al.(2019)Lowenberg, Chodacki, Fenner, Kemp, and Jones]{lowenberg_daniella_2019_3525349}
Daniella Lowenberg, John Chodacki, Martin Fenner, Jennifer Kemp, and Matthew~B. Jones.
\newblock \emph{Open Data Metrics: Lighting the Fire}.
\newblock Zenodo, November 2019.
\newblock URL \url{https://doi.org/10.5281/zenodo.3525349}.

\bibitem[Lowenberg(2022)]{Lowenberg2022Recognizing}
Daniella Lowenberg.
\newblock Recognizing {our} {collective} {responsibility} in the {prioritization} of {open} {data} {metrics}.
\newblock \emph{Harvard Data Science Review}, 4\penalty0 (3), 7 2022.
\newblock URL \url{https://hdsr.mitpress.mit.edu/pub/8vfnfvag}.

\bibitem[Kratz and Strasser(2015)]{kratz2015making}
John~E Kratz and Carly Strasser.
\newblock Making data count.
\newblock \emph{Scientific Data}, 2\penalty0 (1):\penalty0 1--5, 2015.

\bibitem[Fenner et~al.(2018)Fenner, Lowenberg, Jones, Needham, Vieglais, Abrams, Cruse, and Chodacki]{fenner2018code}
Martin Fenner, Daniella Lowenberg, Matt Jones, Paul Needham, Dave Vieglais, Stephen Abrams, Patricia Cruse, and John Chodacki.
\newblock Code of practice for research data usage metrics release 1.
\newblock Technical report, PeerJ Preprints, 2018.

\bibitem[Cousijn et~al.(2019)Cousijn, Feeney, Lowenberg, Presani, and Simons]{cousijn2019bringing}
Helena Cousijn, Patricia Feeney, Daniella Lowenberg, Eleonora Presani, and Natasha Simons.
\newblock Bringing citations and usage metrics together to make data count.
\newblock \emph{Data Science Journal}, 18:\penalty0 9--9, 2019.

\bibitem[Counter(2018)]{counter2018code}
Project Counter.
\newblock {C}ode of {P}ractice for {R}esearch {D}ata, 2018.
\newblock URL \url{https://www.projectcounter.org/code-of-practice-rd-sections/foreword/}.

\bibitem[Burton et~al.(2017)Burton, Fenner, Haak, and Manghi]{scholix2017_1120265}
Adrian Burton, Martin Fenner, Wouter Haak, and Paolo Manghi.
\newblock {Scholix Metadata Schema for Exchange of Scholarly Communication Links}, November 2017.
\newblock URL \url{https://doi.org/10.5281/zenodo.1120265}.

\bibitem[{HuggingFaceFW}(2024)]{huggingfacefw_2024}
{HuggingFaceFW}.
\newblock fineweb (revision af075be), 2024.
\newblock URL \url{https://huggingface.co/datasets/HuggingFaceFW/fineweb}.

\bibitem[Olist and Sionek(2018)]{olist_andr__sionek_2018}
Olist and André Sionek.
\newblock Brazilian e-commerce public dataset by olist, 2018.
\newblock URL \url{https://www.kaggle.com/dsv/195341}.

\bibitem[{DataCite Metadata Working Group}(2023)]{datacite-connection}
{DataCite Metadata Working Group}.
\newblock {Making and Using Connection Metadata}, 2023.
\newblock URL \url{https://support.datacite.org/docs/making-and-using-connection-metadata}.

\bibitem[Logacjov et~al.()Logacjov, Kongsvold, Bach, Bårdstu, and Mork]{harth_779}
Aleksej Logacjov, Atle Kongsvold, Kerstin Bach, Hilde~Bremseth Bårdstu, and Paul~Jarle Mork.
\newblock {HARTH}.
\newblock UCI Machine Learning Repository.
\newblock URL \url{https://doi.org/10.24432/C5NC90}.

\bibitem[Austin et~al.(2017)Austin, Bloom, Dallmeier-Tiessen, Khodiyar, Murphy, Nurnberger, Raymond, Stockhause, Tedds, Vardigan, and Whyte]{austin2017key}
Claire Austin, Theodora Bloom, Sünje Dallmeier-Tiessen, Varsha Khodiyar, Fiona Murphy, Amy Nurnberger, Lisa Raymond, Martina Stockhause, Jonathan Tedds, Mary Vardigan, and Angus Whyte.
\newblock Key components of data publishing: {U}sing current best practices to develop a reference model for data publishing.
\newblock \emph{International Journal on Digital Libraries}, 18\penalty0 (2):\penalty0 77--92, 2017.
\newblock URL \url{https://rdcu.be/cTICG}.

\bibitem[Pasquetto et~al.(2019)Pasquetto, Borgman, and Wofford]{pasquetto2019uses}
Irene~V Pasquetto, Christine~L Borgman, and Morgan~F Wofford.
\newblock Uses and reuses of scientific data: The data creators’ advantage.
\newblock \emph{Harvard Data Science Review}, 2019.

\bibitem[Borgman and Bourne(2022)]{borgman2022why}
Christine~L. Borgman and Philip Bourne.
\newblock Why it takes a village to manage and share data.
\newblock \emph{Harvard Data Science Review}, 4\penalty0 (3), Jul 2022.
\newblock URL \url{https://hdsr.mitpress.mit.edu/pub/wyxni26q}.

\bibitem[Borgman(2019)]{borgman2019lives}
Christine~L. Borgman.
\newblock The lives and after lives of data.
\newblock \emph{Harvard Data Science Review}, 1\penalty0 (1), Jul 2019.
\newblock URL \url{https://hdsr.mitpress.mit.edu/pub/4giycvvj}.

\bibitem[Yadav and Bottou(2019)]{yadav2019cold}
Chhavi Yadav and L{\'e}on Bottou.
\newblock Cold case: The lost {MNIST} digits.
\newblock \emph{Advances in Neural Information Processing Systems}, 32, 2019.

\bibitem[Radin(2017)]{radin2017digital}
Joanna Radin.
\newblock “{D}igital natives”: {H}ow medical and indigenous histories matter for big data.
\newblock \emph{Osiris}, 32\penalty0 (1):\penalty0 43--64, 2017.

\bibitem[Groemping(2019)]{groemping2019south}
Ulrike Groemping.
\newblock South {G}erman credit data: correcting a widely used data set.
\newblock Reports in mathematics, physics, and chemistry, Beuth Hochschule für Technik Berlin, 2019.
\newblock URL \url{http://www1.beuth-hochschule.de/FB_II/reports/Report-2019-004.pdf}.

\bibitem[Gebru et~al.(2021)Gebru, Morgenstern, Vecchione, Vaughan, Wallach, III, and Crawford]{gebru2021datasheets}
Timnit Gebru, Jamie Morgenstern, Briana Vecchione, Jennifer~Wortman Vaughan, Hanna Wallach, Hal~Daum\'{e} III, and Kate Crawford.
\newblock Datasheets for datasets.
\newblock \emph{Communications of the ACM}, 64\penalty0 (12):\penalty0 86–92, 11 2021.
\newblock URL \url{https://doi.org/10.1145/3458723}.

\bibitem[Benjamin et~al.(2019)Benjamin, Gagnon, Rostamzadeh, Pal, Bengio, and Shee]{benjamin2019towards}
Misha Benjamin, Paul Gagnon, Negar Rostamzadeh, Chris Pal, Yoshua Bengio, and Alex Shee.
\newblock Towards standardization of data licenses: The {M}ontreal data license.
\newblock \emph{arXiv preprint arXiv:1903.12262}, 2019.

\bibitem[Rajbahadur et~al.(2021)Rajbahadur, Tuck, Zi, Lin, Chen, Ming, German, et~al.]{rajbahadur2021can}
Gopi~Krishnan Rajbahadur, Erika Tuck, Li~Zi, Dayi Lin, Boyuan Chen, Zhen Ming, Daniel~M German, et~al.
\newblock Can {I} use this publicly available dataset to build commercial {AI} software?--{A} case study on publicly available image datasets.
\newblock \emph{arXiv preprint arXiv:2111.02374}, 2021.

\bibitem[Duan et~al.(2024)Duan, Li, and He]{duan2024modelgo}
Moming Duan, Qinbin Li, and Bingsheng He.
\newblock Modelgo: A practical tool for machine learning license analysis.
\newblock In \emph{Proceedings of the ACM on Web Conference 2024}, WWW '24, page 1158–1169, New York, NY, USA, 2024. Association for Computing Machinery.
\newblock ISBN 9798400701719.
\newblock \doi{10.1145/3589334.3645520}.
\newblock URL \url{https://doi.org/10.1145/3589334.3645520}.

\bibitem[Longpre et~al.(2023)Longpre, Mahari, Chen, Obeng-Marnu, Sileo, Brannon, Muennighoff, Khazam, Kabbara, Perisetla, et~al.]{longpre2023data}
Shayne Longpre, Robert Mahari, Anthony Chen, Naana Obeng-Marnu, Damien Sileo, William Brannon, Niklas Muennighoff, Nathan Khazam, Jad Kabbara, Kartik Perisetla, et~al.
\newblock The data provenance initiative: A large scale audit of dataset licensing \& attribution in {AI}.
\newblock \emph{arXiv preprint arXiv:2310.16787}, 2023.

\bibitem[Northcutt et~al.(2021)Northcutt, Athalye, and Mueller]{northcutt2021pervasive}
Curtis~G Northcutt, Anish Athalye, and Jonas Mueller.
\newblock Pervasive label errors in test sets destabilize machine learning benchmarks.
\newblock In \emph{35th Conference on Neural Information Processing Systems (NeurIPS 2021): Track on Datasets and Benchmarks}, 2021.
\newblock URL \url{https://openreview.net/forum?id=XccDXrDNLek}.

\bibitem[Gururangan et~al.(2018)Gururangan, Swayamdipta, Levy, Schwartz, Bowman, and Smith]{gururangan2018annotation}
Suchin Gururangan, Swabha Swayamdipta, Omer Levy, Roy Schwartz, Samuel Bowman, and Noah~A. Smith.
\newblock Annotation artifacts in natural language inference data.
\newblock In \emph{Proceedings of the 2018 Conference of the North {A}merican Chapter of the Association for Computational Linguistics: Human Language Technologies, Volume 2 (Short Papers)}, pages 107--112, New Orleans, Louisiana, June 2018. Association for Computational Linguistics.
\newblock URL \url{https://aclanthology.org/N18-2017}.

\bibitem[Sasha~Luccioni et~al.(2021)Sasha~Luccioni, Corry, Sridharan, Ananny, Schultz, and Crawford]{sasha2021framework}
Alexandra Sasha~Luccioni, Frances Corry, Hamsini Sridharan, Mike Ananny, Jason Schultz, and Kate Crawford.
\newblock A framework for deprecating datasets: Standardizing documentation, identification, and communication.
\newblock In \emph{Proceedings of the ACM Conference on Fairness, Accountability, and Transparency (FAccT)}. ACM Press, New York, NY, USA, 2021.

\bibitem[Raji and Fried(2020)]{raji2021face}
Inioluwa~Deborah Raji and Genevieve Fried.
\newblock About face: A survey of facial recognition evaluation.
\newblock \emph{Proceedings of the AAAI 2020 Workshop on AI Evaluation}, 2020.

\bibitem[Bandy and Vincent(2021)]{bandy_addressing_2021}
John Bandy and Nicholas Vincent.
\newblock Addressing "documentation debt" in machine learning: {A} retrospective datasheet for {BookCorpus}.
\newblock In \emph{Proceedings of the {Neural} {Information} {Processing} {Systems} {Track} on {Datasets} and {Benchmarks}}, volume~1, 2021.
\newblock URL \url{https://datasets-benchmarks-proceedings.neurips.cc/paper_files/paper/2021/file/54229abfcfa5649e7003b83dd4755294-Paper-round1.pdf}.

\bibitem[Andrews et~al.(2024)Andrews, Zhao, Thong, Modas, Papakyriakopoulos, and Xiang]{andrews2024ethical}
Jerone Andrews, Dora Zhao, William Thong, Apostolos Modas, Orestis Papakyriakopoulos, and Alice Xiang.
\newblock Ethical considerations for responsible data curation.
\newblock \emph{Advances in Neural Information Processing Systems}, 36, 2024.

\bibitem[Birhane and Prabhu(2021)]{birhane2021large}
Abeba Birhane and Vinay~Uday Prabhu.
\newblock Large image datasets: A pyrrhic win for computer vision?
\newblock In \emph{2021 IEEE Winter Conference on Applications of Computer Vision (WACV)}, pages 1536--1546. IEEE, 2021.

\bibitem[Yang et~al.(2020)Yang, Qinami, Fei-Fei, Deng, and Russakovsky]{yang2020towards}
Kaiyu Yang, Klint Qinami, Li~Fei-Fei, Jia Deng, and Olga Russakovsky.
\newblock Towards fairer datasets: {F}iltering and balancing the distribution of the people subtree in the {ImageNet} hierarchy.
\newblock In \emph{Proceedings of the 2020 Conference on Fairness, Accountability, and Transparency}, FAT* '20, page 547–558, New York, NY, USA, 2020. Association for Computing Machinery.
\newblock URL \url{https://doi.org/10.1145/3351095.3375709}.

\bibitem[Buolamwini and Gebru(2018)]{buolamwini2018gender}
Joy Buolamwini and Timnit Gebru.
\newblock Gender shades: Intersectional accuracy disparities in commercial gender classification.
\newblock In \emph{Conference on Fairness, Accountability and Transparency}, pages 77--91. PMLR, 2018.

\bibitem[Merler et~al.(2019)Merler, Ratha, Feris, and Smith]{merler2019diversity}
Michele Merler, Nalini Ratha, Rogerio~S Feris, and John~R Smith.
\newblock Diversity in faces.
\newblock \emph{arXiv preprint arXiv:1901.10436}, 2019.

\bibitem[Park et~al.(2021)Park, Bernstein, Brewer, Kamar, and Morris]{park2021understanding}
Joon~Sung Park, Michael~S. Bernstein, Robin~N. Brewer, Ece Kamar, and Meredith~Ringel Morris.
\newblock Understanding the representation and representativeness of age in {AI} data sets.
\newblock In \emph{Proceedings of the 2021 AAAI/ACM Conference on AI, Ethics, and Society}, AIES '21, page 834–842, New York, NY, USA, 2021. Association for Computing Machinery.
\newblock URL \url{https://doi.org/10.1145/3461702.3462590}.

\bibitem[Carlisle(2019)]{bostonhousing}
M~Carlisle.
\newblock Racist data destruction?, 2019.
\newblock URL \url{https://medium.com/@docintangible/racist-data-destruction-113e3eff54a8}.

\bibitem[Geiger et~al.(2020)Geiger, Yu, Yang, Dai, Qiu, Tang, and Huang]{geiger2020garbage}
R~Stuart Geiger, Kevin Yu, Yanlai Yang, Mindy Dai, Jie Qiu, Rebekah Tang, and Jenny Huang.
\newblock Garbage in, garbage out? {D}o machine learning application papers in social computing report where human-labeled training data comes from?
\newblock In \emph{Proceedings of the 2020 Conference on Fairness, Accountability, and Transparency}, pages 325--336, 2020.

\bibitem[Yang et~al.(2024)Yang, Liang, and Zou]{yang2024navigating}
Xinyu Yang, Weixin Liang, and James Zou.
\newblock Navigating dataset documentations in {AI}: {A} large-scale analysis of dataset cards on {H}ugging {F}ace.
\newblock \emph{ICLR}, 2024.

\bibitem[Hand(2006)]{hand2006classifier}
David~J. Hand.
\newblock Classifier technology and the illusion of progress.
\newblock \emph{Statistical Science}, 21\penalty0 (1):\penalty0 1 -- 14, 2006.
\newblock URL \url{https://doi.org/10.1214/088342306000000060}.

\bibitem[Bender et~al.(2021)Bender, Gebru, McMillan-Major, and Shmitchell]{bender2021dangers}
Emily~M. Bender, Timnit Gebru, Angelina McMillan-Major, and Shmargaret Shmitchell.
\newblock On the dangers of stochastic parrots: Can language models be too big?
\newblock In \emph{Proceedings of the 2021 ACM Conference on Fairness, Accountability, and Transparency (FAccT)}, page 610–623, New York, NY, USA, 2021. Association for Computing Machinery.
\newblock URL \url{https://doi.org/10.1145/3442188.3445922}.

\bibitem[Leavy et~al.(2021)Leavy, Siapera, and O'Sullivan]{leavy2021ethical}
Susan Leavy, Eugenia Siapera, and Barry O'Sullivan.
\newblock Ethical data curation for {AI}: An approach based on feminist epistemology and critical theories of race.
\newblock In \emph{Proceedings of the 2021 AAAI/ACM Conference on AI, Ethics, and Society}, AIES '21, page 695–703, New York, NY, USA, 2021. Association for Computing Machinery.
\newblock URL \url{https://doi.org/10.1145/3461702.3462598}.

\bibitem[Pushkarna et~al.(2022)Pushkarna, Zaldivar, and Kjartansson]{pushkarna2022data}
Mahima Pushkarna, Andrew Zaldivar, and Oddur Kjartansson.
\newblock Data cards: Purposeful and transparent dataset documentation for responsible {AI}.
\newblock In \emph{Proceedings of the 2022 ACM Conference on Fairness, Accountability, and Transparency (FAcct)}, page 1776–1826, New York, NY, USA, 2022. Association for Computing Machinery.
\newblock URL \url{https://doi.org/10.1145/3531146.3533231}.

\bibitem[Holland et~al.(2020)Holland, Hosny, Newman, Joseph, and Chmielinski]{holland2020dataset}
Sarah Holland, Ahmed Hosny, Sarah Newman, Joshua Joseph, and Kasia Chmielinski.
\newblock The dataset nutrition label.
\newblock \emph{Data Protection and Privacy}, 12\penalty0 (12):\penalty0 1, 2020.

\bibitem[Chmielinski et~al.(2022)Chmielinski, Newman, Taylor, Joseph, Thomas, Yurkofsky, and Qiu]{chmielinski2022dataset}
Kasia~S Chmielinski, Sarah Newman, Matt Taylor, Josh Joseph, Kemi Thomas, Jessica Yurkofsky, and Yue~Chelsea Qiu.
\newblock The dataset nutrition label (2nd gen): Leveraging context to mitigate harms in artificial intelligence.
\newblock \emph{arXiv preprint arXiv:2201.03954}, 2022.

\bibitem[Boyd(2021)]{boyd2021datasheets}
Karen~L. Boyd.
\newblock Datasheets for datasets help {ML} engineers notice and understand ethical issues in training data.
\newblock \emph{Proceedings of the ACM on Human-Computer Interaction}, 5\penalty0 (CSCW2), Oct 2021.
\newblock URL \url{https://doi.org/10.1145/3479582}.

\bibitem[Dodge et~al.(2021)Dodge, Sap, Marasovi{\'c}, Agnew, Ilharco, Groeneveld, Mitchell, and Gardner]{dodge2021documenting}
Jesse Dodge, Maarten Sap, Ana Marasovi{\'c}, William Agnew, Gabriel Ilharco, Dirk Groeneveld, Margaret Mitchell, and Matt Gardner.
\newblock Documenting large webtext corpora: A case study on the colossal clean crawled corpus.
\newblock In \emph{Proceedings of the 2021 Conference on Empirical Methods in Natural Language Processing}, pages 1286--1305, 2021.

\bibitem[Biderman and Scheirer(2020)]{biderman2020pitfalls}
Stella Biderman and Walter~J. Scheirer.
\newblock Pitfalls in machine learning research: Reexamining the development cycle.
\newblock In Jessica Zosa~Forde, Francisco Ruiz, Melanie~F. Pradier, and Aaron Schein, editors, \emph{Proceedings on "I Can't Believe It's Not Better!" at NeurIPS Workshops}, volume 137 of \emph{Proceedings of Machine Learning Research}, pages 106--117. PMLR, 12 Dec 2020.
\newblock URL \url{https://proceedings.mlr.press/v137/biderman20a.html}.

\bibitem[Mehrabi et~al.(2021)Mehrabi, Morstatter, Saxena, Lerman, and Galstyan]{mehrabi2022survey}
Ninareh Mehrabi, Fred Morstatter, Nripsuta Saxena, Kristina Lerman, and Aram Galstyan.
\newblock A survey on bias and fairness in machine learning.
\newblock \emph{ACM Computing Surveys}, 54\penalty0 (6), Jul 2021.
\newblock URL \url{https://doi.org/10.1145/3457607}.

\bibitem[Stewart et~al.(2016)Stewart, Andriluka, and Ng]{stewart2016end}
Russell Stewart, Mykhaylo Andriluka, and Andrew~Y Ng.
\newblock End-to-end people detection in crowded scenes.
\newblock In \emph{Proceedings of the IEEE Conference on Computer Vision and Pattern Recognition}, pages 2325--2333, 2016.

\bibitem[Ristani et~al.(2016)Ristani, Solera, Zou, Cucchiara, and Tomasi]{ristani2016performance}
Ergys Ristani, Francesco Solera, Roger Zou, Rita Cucchiara, and Carlo Tomasi.
\newblock Performance measures and a data set for multi-target, multi-camera tracking.
\newblock In \emph{Computer {Vision} – {ECCV} 2016 {Workshops}}, pages 17--35, Cham, 2016. Springer International Publishing.

\bibitem[Founds et~al.(2011)Founds, Orlans, Genevieve, and Watson]{founds2011nist}
Andrew Founds, Nick Orlans, Whiddon Genevieve, and Craig Watson.
\newblock {NIST} special database 32 - multiple encounter dataset {II (MEDS-II)}, Jul 2011.
\newblock URL \url{https://doi.org/10.6028/NIST.IR.7807}.

\bibitem[Phillips et~al.(2003)Phillips, Grother, Micheals, Blackburn, Tabassi, and Bone]{phillips2003face}
P~Phillips, Patrick Grother, Ross Micheals, D~Blackburn, Elham Tabassi, and M~Bone.
\newblock Face recognition vendor test 2002: Evaluation report, 2003-03-01 2003.
\newblock URL \url{https://doi.org/10.6028/NIST.IR.6965}.

\bibitem[Dooley et~al.(2022)Dooley, Wei, Goldstein, and Dickerson]{dooley2022robustness}
Samuel Dooley, George~Z Wei, Tom Goldstein, and John Dickerson.
\newblock Robustness disparities in face detection.
\newblock In \emph{Advances in Neural Information Processing Systems}, volume~35, pages 38245--38259. Curran Associates, Inc., 2022.

\bibitem[Zhu et~al.(2023)Zhu, Huang, Deng, Ye, Huang, Chen, Zhu, Yang, Du, Lu, and Zhou]{zhu2023webface}
Zheng Zhu, Guan Huang, Jiankang Deng, Yun Ye, Junjie Huang, Xinze Chen, Jiagang Zhu, Tian Yang, Dalong Du, Jiwen Lu, and Jie Zhou.
\newblock {WebFace260M}: A benchmark for million-scale deep face recognition.
\newblock \emph{IEEE Transactions on Pattern Analysis and Machine Intelligence}, 45\penalty0 (2):\penalty0 2627--2644, 2023.
\newblock URL \url{https://doi.org/10.1109/TPAMI.2022.3169734}.

\bibitem[Rangineni(2023)]{rangineni2023analysis}
Sandeep Rangineni.
\newblock An analysis of data quality requirements for machine learning development pipelines frameworks.
\newblock \emph{International Journal of Computer Trends and Technology}, 71\penalty0 (9):\penalty0 16--27, 2023.

\bibitem[Tsipras et~al.(2020)Tsipras, Santurkar, Engstrom, Ilyas, and Madry]{tsipras2020imagenet}
Dimitris Tsipras, Shibani Santurkar, Logan Engstrom, Andrew Ilyas, and Aleksander Madry.
\newblock From {ImageNet} to image classification: Contextualizing progress on benchmarks.
\newblock In \emph{Proceedings of the 2020 International Conference on Machine Learning}, pages 9625--9635. PMLR, 2020.

\bibitem[Sen et~al.(2015)Sen, Giesel, Gold, Hillmann, Lesicko, Naden, Russell, Wang, and Hecht]{shilad2015turkers}
Shilad Sen, Margaret~E. Giesel, Rebecca Gold, Benjamin Hillmann, Matt Lesicko, Samuel Naden, Jesse Russell, Zixiao~(Ken) Wang, and Brent Hecht.
\newblock Turkers, scholars, "arafat" and "peace": Cultural communities and algorithmic gold standards.
\newblock In \emph{Proceedings of the 18th ACM Conference on Computer Supported Cooperative Work \& Social Computing}, CSCW '15, page 826–838, New York, NY, USA, 2015. Association for Computing Machinery.
\newblock URL \url{https://doi.org/10.1145/2675133.2675285}.

\bibitem[Bowman and Dahl(2021)]{bowman2021fix}
Samuel~R. Bowman and George Dahl.
\newblock What will it take to fix benchmarking in natural language understanding?
\newblock In \emph{Proceedings of the 2021 Conference of the North American Chapter of the Association for Computational Linguistics: Human Language Technologies}, pages 4843--4855, Online, June 2021. Association for Computational Linguistics.
\newblock URL \url{https://doi.org/10.18653/v1/2021.naacl-main.385}.

\bibitem[Daneshjou et~al.(2022)Daneshjou, Vodrahalli, Novoa, Jenkins, Liang, Rotemberg, Ko, Swetter, Bailey, Gevaert, Mukherjee, Phung, Yekrang, Fong, Sahasrabudhe, Allerup, Okata-Karigane, Zou, and Chiou]{daneshjou2022disparities}
R.~Daneshjou, K.~Vodrahalli, R.~A. Novoa, M.~Jenkins, W.~Liang, V.~Rotemberg, J.~Ko, S.~M. Swetter, E.~E. Bailey, O.~Gevaert, P.~Mukherjee, M.~Phung, K.~Yekrang, B.~Fong, R.~Sahasrabudhe, J.~A.~C. Allerup, U.~Okata-Karigane, J.~Zou, and A.~S. Chiou.
\newblock Disparities in dermatology {AI} performance on a diverse, curated clinical image set.
\newblock \emph{Science Advances}, 8\penalty0 (32), 2022.
\newblock URL \url{https://doi.org/10.1126/sciadv.abq6147}.

\bibitem[Sörqvist et~al.(2011)Sörqvist, Langeborg, and Eriksson]{sorqvist2011women}
Patrik Sörqvist, Linda Langeborg, and Mårten Eriksson.
\newblock Women assimilate across gender, men don't: The role of gender to the own-anchor effect in age, height, and weight estimates.
\newblock \emph{Journal of Applied Social Psychology}, 41\penalty0 (7):\penalty0 1733--1748, 2011.
\newblock URL \url{https://doi.org/10.1111/j.1559-1816.2011.00774.x}.

\bibitem[Hofmann(1994)]{statlog_german_credit_data_144}
Hans Hofmann.
\newblock {Statlog (German Credit Data)}.
\newblock UCI Machine Learning Repository, 1994.
\newblock URL \url{https://doi.org/10.24432/C5NC77}.

\bibitem[H{\"a}ußler(1979)]{haubetaler1979empirische}
WM~H{\"a}ußler.
\newblock Empirische ergebnisse zu diskriminationsverfahren bei kreditscoringsystemen.
\newblock \emph{Zeitschrift f{\"u}r Operations Research}, 23:\penalty0 B191--B210, 1979.

\bibitem[H{\"a}ußler(1981)]{haubetaler1981methoden}
WM~H{\"a}ußler.
\newblock Methoden der punktebewertung f{\"u}r kreditscoringsysteme.
\newblock \emph{Zeitschrift f{\"u}r Operations Research}, 25:\penalty0 B79--B94, 1981.

\bibitem[Fahrmeir and Hamerle(1981)]{fahrmeir1981kategoriale}
Ludwig Fahrmeir and Alfred Hamerle.
\newblock Kategoriale regression in der betrieblichen planung.
\newblock \emph{Zeitschrift f{\"u}r Operations Research}, 25:\penalty0 B63--B78, 1981.

\bibitem[Fahrmeir and Hamerle(1984)]{fahrmeir1984multivariate}
Ludwig Fahrmeir and Alfred Hamerle.
\newblock \emph{Multivariate Statistische Verfahren}.
\newblock de Gruyter, 1984.

\bibitem[Srikumar et~al.(2022)Srikumar, Finlay, Abuhamad, Ashurst, Campbell, Campbell-Ratcliffe, Hongo, Jordan, Lindley, Ovadya, et~al.]{srikumar2022advancing}
Madhulika Srikumar, Rebecca Finlay, Grace Abuhamad, Carolyn Ashurst, Rosie Campbell, Emily Campbell-Ratcliffe, Hudson Hongo, Sara~R Jordan, Joseph Lindley, Aviv Ovadya, et~al.
\newblock Advancing ethics review practices in {AI} research.
\newblock \emph{Nature Machine Intelligence}, 4\penalty0 (12):\penalty0 1061--1064, 2022.

\bibitem[Fisher(1988)]{misc_iris_53}
R.A. Fisher.
\newblock {Iris}.
\newblock UCI Machine Learning Repository, 1988.
\newblock URL \url{https://doi.org/10.24432/C56C76}.

\bibitem[Bezdek et~al.(1999)Bezdek, Keller, Krishnapuram, Kuncheva, and Pal]{bezdek1999will}
James~C Bezdek, James~M Keller, Raghu Krishnapuram, Ludmila~I Kuncheva, and Nikhil~R Pal.
\newblock Will the real iris data please stand up?
\newblock \emph{IEEE Transactions on Fuzzy Systems}, 7\penalty0 (3):\penalty0 368--369, 1999.

\bibitem[Sawchuk and Khair(2021)]{Sawchuk2021qy}
Sandra~L Sawchuk and Shahira Khair.
\newblock Computational reproducibility: A practical framework for data curators.
\newblock \emph{Journal of eScience Librarianship}, 10\penalty0 (3):\penalty0 7, 2021.

\bibitem[Sculley et~al.(2015)Sculley, Holt, Golovin, Davydov, Phillips, Ebner, Chaudhary, Young, Crespo, and Dennison]{sculley2015hidden}
D.~Sculley, Gary Holt, Daniel Golovin, Eugene Davydov, Todd Phillips, Dietmar Ebner, Vinay Chaudhary, Michael Young, Jean-Fran\c{c}ois Crespo, and Dan Dennison.
\newblock Hidden technical debt in machine learning systems.
\newblock In \emph{Advances in Neural Information Processing Systems}, volume~28. Curran Associates, Inc., 2015.
\newblock URL \url{https://proceedings.neurips.cc/paper_files/paper/2015/file/86df7dcfd896fcaf2674f757a2463eba-Paper.pdf}.

\bibitem[Sheridan et~al.(2021)Sheridan, Dellureficio, Ratajeski, Mannheimer, and Wheeler]{Sheridan2021xu}
Helenmary Sheridan, Anthony~J Dellureficio, Melissa~A Ratajeski, Sara Mannheimer, and Terrie~R Wheeler.
\newblock Data curation through catalogs: A repository-independent model for data discovery.
\newblock \emph{Journal of eScience Librarianship}, 10\penalty0 (3):\penalty0 4, 2021.

\bibitem[Forsyth(1990)]{misc_bupa_60}
Richard Forsyth.
\newblock {Liver Disorders}.
\newblock UCI Machine Learning Repository, 1990.
\newblock URL \url{https://doi.org/10.24432/C54G67}.

\bibitem[McDermott and Forsyth(2016)]{MCDERMOTT201641}
James McDermott and Richard~S. Forsyth.
\newblock Diagnosing a disorder in a classification benchmark.
\newblock \emph{Pattern Recognition Letters}, 73:\penalty0 41--43, 2016.
\newblock URL \url{https://doi.org/10.1016/j.patrec.2016.01.004}.

\bibitem[Akhtar et~al.(2024)Akhtar, Benjelloun, Conforti, Gijsbers, Giner-Miguelez, Jain, Kuchnik, Lhoest, Marcenac, Maskey, et~al.]{akhtar2024croissant}
Mubashara Akhtar, Omar Benjelloun, Costanza Conforti, Pieter Gijsbers, Joan Giner-Miguelez, Nitisha Jain, Michael Kuchnik, Quentin Lhoest, Pierre Marcenac, Manil Maskey, et~al.
\newblock Croissant: A metadata format for {ML}-ready datasets.
\newblock In \emph{Proceedings of the Eighth Workshop on Data Management for End-to-End Machine Learning}, pages 1--6, 2024.

\bibitem[Raff(2019)]{raff2019reproducible}
Edward Raff.
\newblock A step toward quantifying independently reproducible machine learning research.
\newblock In \emph{Advances in Neural Information Processing Systems}, volume~32. Curran Associates, Inc., 2019.
\newblock URL \url{https://proceedings.neurips.cc/paper_files/paper/2019/file/c429429bf1f2af051f2021dc92a8ebea-Paper.pdf}.

\bibitem[Henderson et~al.(2018)Henderson, Islam, Bachman, Pineau, Precup, and Meger]{henderson2018deep}
Peter Henderson, Riashat Islam, Philip Bachman, Joelle Pineau, Doina Precup, and David Meger.
\newblock Deep reinforcement learning that matters.
\newblock In \emph{Proceedings of the AAAI Conference on Artificial Intelligence}, volume~32. AAAI Press, Menlo Park, 2018.
\newblock URL \url{https://doi.org/10.1609/aaai.v32i1.11694}.

\bibitem[Friedland(2024)]{Friedland2024}
Gerald Friedland.
\newblock \emph{Repeatability and Reproducibility}, pages 201--208.
\newblock Springer International Publishing, Cham, 2024.
\newblock URL \url{https://doi.org/10.1007/978-3-031-39477-5_15}.

\bibitem[Maffia et~al.(2023)Maffia, Burkhart, Ciorba, and Resch]{maffia2023benchmarking}
Antonio Maffia, Helmar Burkhart, Florina~M. Ciorba, and Michael Resch.
\newblock \emph{On Benchmarking of Deep Learning Systems: Software Engineering Issues and Reproducibility Challenges}.
\newblock Philosophisch-Naturwissenschaftliche Fakult{\"a}t der Universit{\"a}t Basel, 2023.
\newblock URL \url{https://books.google.com/books?id=rRw50AEACAAJ}.

\bibitem[Gundersen et~al.(2022)Gundersen, Coakley, Kirkpatrick, and Gil]{gundersen2022sources}
Odd~Erik Gundersen, Kevin Coakley, Christine Kirkpatrick, and Yolanda Gil.
\newblock Sources of irreproducibility in machine learning: A review.
\newblock \emph{arXiv preprint arXiv:2204.07610}, 2022.

\bibitem[Li et~al.(2023)Li, Qi, Liu, Di, Liu, Pei, Yi, and Zhou]{li2023trustworthy}
Bo~Li, Peng Qi, Bo~Liu, Shuai Di, Jingen Liu, Jiquan Pei, Jinfeng Yi, and Bowen Zhou.
\newblock Trustworthy {AI}: From principles to practices.
\newblock \emph{ACM Computing Surveys}, 55\penalty0 (9), Jan 2023.
\newblock URL \url{https://doi.org/10.1145/3555803}.

\bibitem[Tatman et~al.(2018)Tatman, Vanderplas, and Dane]{Tatman2018APT}
Rachael Tatman, Jake Vanderplas, and Sohier Dane.
\newblock A practical taxonomy of reproducibility for machine learning research.
\newblock In \emph{2nd Reproducibility in Machine Learning Workshop}, 2018.

\bibitem[Hulsebos et~al.(2024)Hulsebos, Lin, Shankar, and Parameswaran]{hulsebos2024longer}
Madelon Hulsebos, Wenjing Lin, Shreya Shankar, and Aditya~G. Parameswaran.
\newblock “{I}t took longer than {I} was expecting:” {W}hy is dataset search still so hard?
\newblock Preprint at \url{https://www.madelonhulsebos.com/assets/dataset_search_survey.pdf}., 2024.

\bibitem[Giner-Miguelez et~al.(2023)Giner-Miguelez, Gómez, and Cabot]{giner2023domain}
Joan Giner-Miguelez, Abel Gómez, and Jordi Cabot.
\newblock A domain-specific language for describing machine learning datasets.
\newblock \emph{Journal of Computer Languages}, 76:\penalty0 101209, 2023.
\newblock URL \url{https://doi.org/10.1016/j.cola.2023.101209}.

\bibitem[Lin et~al.(2020)Lin, Crabtree, Dillo, Downs, Edmunds, Giaretta, De~Giusti, L’Hours, Hugo, Jenkyns, et~al.]{lin2020trust}
Dawei Lin, Jonathan Crabtree, Ingrid Dillo, Robert~R Downs, Rorie Edmunds, David Giaretta, Marisa De~Giusti, Herv{\'e} L’Hours, Wim Hugo, Reyna Jenkyns, et~al.
\newblock The {TRUST} principles for digital repositories.
\newblock \emph{Scientific Data}, 7\penalty0 (1):\penalty0 144, 2020.

\bibitem[NIH(2022)]{nih2022selecting}
NIH.
\newblock {S}electing a data repository, 2022.
\newblock URL \url{https://sharing.nih.gov/data-management-and-sharing-policy/sharing-scientific-data/selecting-a-data-repository}.

\bibitem[{CCSDS}(2012)]{oais}
{CCSDS}.
\newblock Reference model for an open archival information system ({OAIS}), 2012.
\newblock URL \url{http://www.oais.info/}.

\bibitem[{National Artificial Intelligence Research Resource Task Force}(2023)]{nair2023report}
{National Artificial Intelligence Research Resource Task Force}.
\newblock Strengthening and democratizing the {U.S.} artificial intelligence innovation ecosystem: {A}n implementation plan for a national artificial intelligence research resource, 2023.
\newblock URL \url{https://www.ai.gov/wp-content/uploads/2023/01/NAIRR-TF-Final-Report-2023.pdf}.

\bibitem[Elfer et~al.(2024)Elfer, Gardecki, Garcia, Ly, and et~al.]{elfer2024reproducible}
Katherine Elfer, Emma Gardecki, Victor Garcia, Amy Ly, and Hytopoulous et~al.
\newblock Reproducible reporting of the collection and evaluation of annotations for artificial intelligence models.
\newblock \emph{Modern Pathology}, 37\penalty0 (4):\penalty0 100439, 2024.
\newblock URL \url{https://doi.org/10.1016/j.modpat.2024.100439}.

\bibitem[Pawlik et~al.(2019)Pawlik, H{\"u}tter, Kocher, Mann, and Augsten]{pawlik2019link}
Mateusz Pawlik, Thomas H{\"u}tter, Daniel Kocher, Willi Mann, and Nikolaus Augsten.
\newblock A link is not enough--reproducibility of data.
\newblock \emph{Datenbank-Spektrum}, 19:\penalty0 107--115, 2019.

\bibitem[Welty et~al.(2019)Welty, Paritosh, and Aroyo]{welty2019metrology}
Chris Welty, Praveen Paritosh, and Lora Aroyo.
\newblock Metrology for {AI}: From benchmarks to instruments.
\newblock \emph{arXiv preprint arXiv:1911.01875}, 2019.

\bibitem[Shavit and Ellison(2017)]{shavit2017stepping}
Ayelet Shavit and Aaron~M Ellison.
\newblock \emph{Stepping in the same river twice: Replication in biological research}.
\newblock Yale University Press, 2017.

\bibitem[Stoyanovich and Howe(2019)]{stoyanovich2019nutritional}
Julia Stoyanovich and Bill Howe.
\newblock Nutritional labels for data and models.
\newblock \emph{A Quarterly bulletin of the Computer Society of the IEEE Technical Committee on Data Engineering}, 42\penalty0 (3), 2019.

\bibitem[Bender and Friedman(2018)]{bender2018data}
Emily~M. Bender and Batya Friedman.
\newblock Data statements for natural language processing: Toward mitigating system bias and enabling better science.
\newblock \emph{Transactions of the Association for Computational Linguistics}, 6:\penalty0 587--604, 2018.
\newblock URL \url{https://doi.org/10.1162/tacl_a_00041}.

\bibitem[Xu et~al.(2022)Xu, Watts, Bankston, and Sare]{Xu2022cn}
Zhihong Xu, John Watts, Sarah Bankston, and Laura Sare.
\newblock Depositing data: A usability study of the {T}exas data repository.
\newblock \emph{Journal of eScience Librarianship}, 11\penalty0 (1):\penalty0 6, 2022.

\bibitem[{U.S. Bureau of Economic Analysis}(2022)]{acdeb2022report}
{U.S. Bureau of Economic Analysis}.
\newblock Advisory committee on data for evidence building: {Y}ear 2 report.
\newblock \url{https://www.bea.gov/system/files/2022-10/acdeb-year-2-report.pdf}, 2022.

\bibitem[Bouthillier et~al.(2021)Bouthillier, Delaunay, Bronzi, Trofimov, Nichyporuk, Szeto, Mohammadi~Sepahvand, Raff, Madan, Voleti, et~al.]{bouthillier2021accounting}
Xavier Bouthillier, Pierre Delaunay, Mirko Bronzi, Assya Trofimov, Brennan Nichyporuk, Justin Szeto, Nazanin Mohammadi~Sepahvand, Edward Raff, Kanika Madan, Vikram Voleti, et~al.
\newblock Accounting for variance in machine learning benchmarks.
\newblock \emph{Proceedings of Machine Learning and Systems}, 3:\penalty0 747--769, 2021.

\bibitem[Pineau et~al.(2021)Pineau, Vincent-Lamarre, Sinha, Larivi\`{e}re, Beygelzimer, d'Alch\'{e} Buc, Fox, and Larochelle]{pineau2021improving}
Joelle Pineau, Philippe Vincent-Lamarre, Koustuv Sinha, Vincent Larivi\`{e}re, Alina Beygelzimer, Florence d'Alch\'{e} Buc, Emily Fox, and Hugo Larochelle.
\newblock Improving reproducibility in machine learning research (a report from the {N}eur{IPS} 2019 reproducibility program).
\newblock \emph{Journal of Machine Learning Research}, 22\penalty0 (164):\penalty0 1--20, Jan 2021.
\newblock URL \url{https://dl.acm.org/doi/10.5555/3546258.3546422}.

\bibitem[Kang et~al.(2023)Kang, Kang, and Jang]{kang2023papers}
Donghyun Kang, TaeYoung Kang, and Junkyu Jang.
\newblock Papers with code or without code? {Impact of GitHub} repository usability on the diffusion of machine learning research.
\newblock \emph{Information Processing \& Management}, 60\penalty0 (6):\penalty0 103477, 2023.
\newblock URL \url{https://doi.org/10.1016/j.ipm.2023.103477}.

\bibitem[Gundersen(2021)]{gundersen2021fundamental}
Odd~Erik Gundersen.
\newblock The fundamental principles of reproducibility.
\newblock \emph{Philosophical Transactions of the Royal Society A}, 379\penalty0 (2197):\penalty0 20200210, 2021.

\bibitem[Laurinavichyute et~al.(2022)Laurinavichyute, Yadav, and Vasishth]{laurinavichyute2022share}
Anna Laurinavichyute, Himanshu Yadav, and Shravan Vasishth.
\newblock Share the code, not just the data: A case study of the reproducibility of articles published in the journal of memory and language under the open data policy.
\newblock \emph{Journal of Memory and Language}, 125:\penalty0 104332, 2022.

\bibitem[Wang et~al.(2021)Wang, Wang, Drozdal, Liu, Park, Oney, and Brooks]{wang2021kaggle}
April~Yi Wang, Dakuo Wang, Jaimie Drozdal, Xuye Liu, Soya Park, Steve Oney, and Christopher Brooks.
\newblock What makes a well-documented notebook? {A} case study of data scientists’ documentation practices in {Kaggle}.
\newblock In \emph{Extended Abstracts of the 2021 CHI Conference on Human Factors in Computing Systems}, CHI EA '21, New York, NY, USA, 2021. Association for Computing Machinery.
\newblock URL \url{https://doi.org/10.1145/3411763.3451617}.

\bibitem[Leipzig et~al.(2021)Leipzig, N{\"u}st, Hoyt, Ram, and Greenberg]{leipzig2021role}
Jeremy Leipzig, Daniel N{\"u}st, Charles~Tapley Hoyt, Karthik Ram, and Jane Greenberg.
\newblock The role of metadata in reproducible computational research.
\newblock \emph{Patterns}, 2\penalty0 (9):\penalty0 100322, 2021.

\bibitem[Teney et~al.(2020)Teney, Abbasnejad, Kafle, Shrestha, Kanan, and Van Den~Hengel]{teney2020value}
Damien Teney, Ehsan Abbasnejad, Kushal Kafle, Robik Shrestha, Christopher Kanan, and Anton Van Den~Hengel.
\newblock On the value of out-of-distribution testing: An example of {G}oodhart's law.
\newblock \emph{Advances in Neural Information Processing Systems}, 33:\penalty0 407--417, 2020.

\bibitem[Thomas and Uminsky(2022)]{thomas2022reliance}
Rachel~L Thomas and David Uminsky.
\newblock Reliance on metrics is a fundamental challenge for {AI}.
\newblock \emph{Patterns}, 3\penalty0 (5), 2022.

\bibitem[Everingham et~al.(2015)Everingham, Eslami, Van~Gool, Williams, Winn, and Zisserman]{everingham2015pascal}
Mark Everingham, SM~Ali Eslami, Luc Van~Gool, Christopher~KI Williams, John Winn, and Andrew Zisserman.
\newblock The pascal visual object classes challenge: A retrospective.
\newblock \emph{International Journal of Computer Vision}, 111:\penalty0 98--136, 2015.

\bibitem[Fern{{\'a}}ndez-Delgado et~al.(2014)Fern{{\'a}}ndez-Delgado, Cernadas, Barro, and Amorim]{fernandez2014hundreds}
Manuel Fern{{\'a}}ndez-Delgado, Eva Cernadas, Sen{{\'e}}n Barro, and Dinani Amorim.
\newblock Do we need hundreds of classifiers to solve real world classification problems?
\newblock \emph{Journal of Machine Learning Research}, 15\penalty0 (90):\penalty0 3133--3181, 2014.
\newblock URL \url{http://jmlr.org/papers/v15/delgado14a.html}.

\bibitem[Lin et~al.(2021)Lin, Campos, Craswell, Mitra, and Yilmaz]{lin2021significant}
Jimmy Lin, Daniel Campos, Nick Craswell, Bhaskar Mitra, and Emine Yilmaz.
\newblock Significant improvements over the state of the art? {A} case study of the ms marco document ranking leaderboard.
\newblock In \emph{Proceedings of the 44th International ACM SIGIR Conference on Research and Development in Information Retrieval}, pages 2283--2287, 2021.

\bibitem[Burnell et~al.(2023)Burnell, Schellaert, Burden, Ullman, Martinez-Plumed, Tenenbaum, Rutar, Cheke, Sohl-Dickstein, Mitchell, Kiela, Shanahan, Voorhees, Cohn, Leibo, and Hernandez-Orallo]{burnell2023rethink}
Ryan Burnell, Wout Schellaert, John Burden, Tomer~D. Ullman, Fernando Martinez-Plumed, Joshua~B. Tenenbaum, Danaja Rutar, Lucy~G. Cheke, Jascha Sohl-Dickstein, Melanie Mitchell, Douwe Kiela, Murray Shanahan, Ellen~M. Voorhees, Anthony~G. Cohn, Joel~Z. Leibo, and Jose Hernandez-Orallo.
\newblock Rethink reporting of evaluation results in {AI}.
\newblock \emph{Science}, 380\penalty0 (6641):\penalty0 136--138, 2023.
\newblock URL \url{https://www.science.org/doi/abs/10.1126/science.adf6369}.

\bibitem[Ribeiro et~al.(2020)Ribeiro, Wu, Guestrin, and Singh]{ribeiro2020beyond}
Marco~Tulio Ribeiro, Tongshuang Wu, Carlos Guestrin, and Sameer Singh.
\newblock Beyond accuracy: Behavioral testing of {NLP} models with {CheckList}.
\newblock In \emph{Proceedings of the 58th Annual Meeting of the Association for Computational Linguistics}. Association for Computational Linguistics, 2020.

\bibitem[Ethayarajh and Jurafsky(2020)]{ethayarajh2020utility}
Kawin Ethayarajh and Dan Jurafsky.
\newblock Utility is in the eye of the user: A critique of {NLP} leaderboards.
\newblock In \emph{Proceedings of the 2020 Conference on Empirical Methods in Natural Language Processing (EMNLP)}, pages 4846--4853, Online, November 2020. Association for Computational Linguistics.
\newblock URL \url{https://aclanthology.org/2020.emnlp-main.393}.

\bibitem[Lipton and Steinhardt(2019)]{lipton2019troubling}
Zachary~C. Lipton and Jacob Steinhardt.
\newblock Troubling trends in machine learning scholarship: Some {ML} papers suffer from flaws that could mislead the public and stymie future research.
\newblock \emph{Queue}, 17\penalty0 (1):\penalty0 45–77, Feb 2019.
\newblock URL \url{https://doi.org/10.1145/3317287.3328534}.

\bibitem[Wu et~al.(2019)Wu, Ribeiro, Heer, and Weld]{wu2019errudite}
Tongshuang Wu, Marco~Tulio Ribeiro, Jeffrey Heer, and Daniel Weld.
\newblock {E}rrudite: Scalable, reproducible, and testable error analysis.
\newblock In \emph{Proceedings of the 57th Annual Meeting of the Association for Computational Linguistics}, pages 747--763, Florence, Italy, July 2019. Association for Computational Linguistics.
\newblock URL \url{https://aclanthology.org/P19-1073}.

\bibitem[Tötsch and Hoffmann(2021)]{totsch2021}
Niklas Tötsch and Daniel Hoffmann.
\newblock Classifier uncertainty: evidence, potential impact, and probabilistic treatment.
\newblock \emph{PeerJ Computer Science}, 7:\penalty0 e398, 2021.
\newblock URL \url{https://doi.org/10.7717/peerj-cs.398}.

\bibitem[Graham et~al.(2014)Graham, Mathur, and Baldwin]{graham2014randomized}
Yvette Graham, Nitika Mathur, and Timothy Baldwin.
\newblock Randomized significance tests in machine translation.
\newblock In \emph{Proceedings of the Ninth Workshop on Statistical Machine Translation}, pages 266--274, Baltimore, Maryland, USA, June 2014. Association for Computational Linguistics.
\newblock URL \url{https://aclanthology.org/W14-3333}.

\bibitem[Dror et~al.(2017)Dror, Baumer, Bogomolov, and Reichart]{dror2017replicability}
Rotem Dror, Gili Baumer, Marina Bogomolov, and Roi Reichart.
\newblock Replicability analysis for natural language processing: Testing significance with multiple datasets.
\newblock \emph{Transactions of the Association for Computational Linguistics}, 5:\penalty0 471--486, 2017.
\newblock URL \url{https://aclanthology.org/Q17-1033}.

\bibitem[Wein et~al.(2023)Wein, Homan, Aroyo, and Welty]{wein2023follow}
Shira Wein, Christopher Homan, Lora Aroyo, and Chris Welty.
\newblock Follow the leader(board) with confidence: Estimating p-values from a single test set with item and response variance.
\newblock In \emph{Findings of the Association for Computational Linguistics: ACL 2023}, pages 3138--3161, Toronto, Canada, July 2023. Association for Computational Linguistics.
\newblock URL \url{https://aclanthology.org/2023.findings-acl.196}.

\bibitem[M{\o}llersen and Holsb{\o}(2023)]{Mllersen2023WhatIT}
Kajsa M{\o}llersen and Einar Holsb{\o}.
\newblock What is the state of the art? {A}ccounting for multiplicity in machine learning benchmark performance.
\newblock \emph{arXiv preprint arXiv:2303.07272}, 2023.

\bibitem[Recht et~al.(2019)Recht, Roelofs, Schmidt, and Shankar]{pmlrrecht19a}
Benjamin Recht, Rebecca Roelofs, Ludwig Schmidt, and Vaishaal Shankar.
\newblock Do {I}mage{N}et classifiers generalize to {I}mage{N}et?
\newblock In \emph{Proceedings of the 2019 International Conference on Machine Learning}, volume~97 of \emph{Proceedings of Machine Learning Research}, pages 5389--5400. PMLR, 09--15 Jun 2019.
\newblock URL \url{https://proceedings.mlr.press/v97/recht19a.html}.

\bibitem[Wang et~al.(2019)Wang, Singh, Michael, Hill, Levy, and Bowman]{wang2019glue}
Alex Wang, Amanpreet Singh, Julian Michael, Felix Hill, Omer Levy, and Samuel~R Bowman.
\newblock Glue: A multi-task benchmark and analysis platform for natural language understanding.
\newblock In \emph{7th International Conference on Learning Representations, ICLR 2019}, 2019.

\bibitem[McCoy et~al.(2019)McCoy, Pavlick, and Linzen]{mccoy2019right}
Tom McCoy, Ellie Pavlick, and Tal Linzen.
\newblock Right for the wrong reasons: Diagnosing syntactic heuristics in natural language inference.
\newblock In \emph{Proceedings of the 57th Annual Meeting of the Association for Computational Linguistics}, pages 3428--3448, Florence, Italy, July 2019. Association for Computational Linguistics.
\newblock URL \url{https://aclanthology.org/P19-1334}.

\bibitem[Schlangen(2021)]{schlangen2021targeting}
David Schlangen.
\newblock Targeting the benchmark: On methodology in current natural language processing research.
\newblock In \emph{Proceedings of the 59th Annual Meeting of the Association for Computational Linguistics and the 11th International Joint Conference on Natural Language Processing (Volume 2: Short Papers)}, Online, August 2021. Association for Computational Linguistics.
\newblock URL \url{https://aclanthology.org/2021.acl-short.85}.

\bibitem[Dwork et~al.(2015)Dwork, Feldman, Hardt, Pitassi, Reingold, and Roth]{dwork2015preserving}
Cynthia Dwork, Vitaly Feldman, Moritz Hardt, Toniann Pitassi, Omer Reingold, and Aaron~Leon Roth.
\newblock Preserving statistical validity in adaptive data analysis.
\newblock In \emph{Proceedings of the Forty-Seventh Annual ACM Symposium on Theory of Computing}, pages 117--126, 2015.

\bibitem[Blum and Hardt(2015)]{blum2015ladder}
Avrim Blum and Moritz Hardt.
\newblock The ladder: A reliable leaderboard for machine learning competitions.
\newblock In \emph{Proceedings of the 2015 International Conference on Machine Learning}, pages 1006--1014. PMLR, 2015.

\bibitem[Roelofs et~al.(2019)Roelofs, Shankar, Recht, Fridovich-Keil, Hardt, Miller, and Schmidt]{roelofs2019meta}
Rebecca Roelofs, Vaishaal Shankar, Benjamin Recht, Sara Fridovich-Keil, Moritz Hardt, John Miller, and Ludwig Schmidt.
\newblock A meta-analysis of overfitting in machine learning.
\newblock In \emph{Advances in Neural Information Processing Systems}, volume~32. Curran Associates, Inc., 2019.
\newblock URL \url{https://proceedings.neurips.cc/paper_files/paper/2019/file/ee39e503b6bedf0c98c388b7e8589aca-Paper.pdf}.

\bibitem[Koh et~al.(2021)Koh, Sagawa, Marklund, Xie, Zhang, Balsubramani, Hu, Yasunaga, Phillips, Gao, Lee, David, Stavness, Guo, Earnshaw, Haque, Beery, Leskovec, Kundaje, Pierson, Levine, Finn, and Liang]{koh2021wilds}
Pang~Wei Koh, Shiori Sagawa, Henrik Marklund, Sang~Michael Xie, Marvin Zhang, Akshay Balsubramani, Weihua Hu, Michihiro Yasunaga, Richard~Lanas Phillips, Irena Gao, Tony Lee, Etienne David, Ian Stavness, Wei Guo, Berton Earnshaw, Imran Haque, Sara~M Beery, Jure Leskovec, Anshul Kundaje, Emma Pierson, Sergey Levine, Chelsea Finn, and Percy Liang.
\newblock Wilds: A benchmark of in-the-wild distribution shifts.
\newblock In \emph{Proceedings of the 2021 International Conference on Machine Learning}, volume 139 of \emph{Proceedings of Machine Learning Research}, pages 5637--5664. PMLR, 18--24 Jul 2021.
\newblock URL \url{https://proceedings.mlr.press/v139/koh21a.html}.

\bibitem[{DataCite Metadata Working Group}(2021)]{datacite}
{DataCite Metadata Working Group}.
\newblock Datacite metadata schema documentation for the publication and citation of research data and other research outputs, 2021.
\newblock URL \url{https://doi.org/10.14454/3w3z-sa82}.

\end{thebibliography}

\section*{Checklist}

\begin{enumerate}

\item For all authors...
\begin{enumerate}
  \item Do the main claims made in the abstract and introduction accurately reflect the paper's contributions and scope?
    \answerYes{}
  \item Did you describe the limitations of your work?
    \answerYes{}
  \item Did you discuss any potential negative societal impacts of your work?
    \answerNo{}
  \item Have you read the ethics review guidelines and ensured that your paper conforms to them?
    \answerYes{}
\end{enumerate}

\item If you are including theoretical results...
\begin{enumerate}
  \item Did you state the full set of assumptions of all theoretical results?
    \answerNA{}
	\item Did you include complete proofs of all theoretical results?
    \answerNA{}
\end{enumerate}

\item If you ran experiments (e.g. for benchmarks)...
\begin{enumerate}
  \item Did you include the code, data, and instructions needed to reproduce the main experimental results (either in the supplemental material or as a URL)?
    \answerNA{}
  \item Did you specify all the training details (e.g., data splits, hyperparameters, how they were chosen)?
    \answerNA{}
	\item Did you report error bars (e.g., with respect to the random seed after running experiments multiple times)?
    \answerNA{}
	\item Did you include the total amount of compute and the type of resources used (e.g., type of GPUs, internal cluster, or cloud provider)?
    \answerNA{}
\end{enumerate}

\item If you are using existing assets (e.g., code, data, models) or curating/releasing new assets...
\begin{enumerate}
  \item If your work uses existing assets, did you cite the creators?
    \answerNA{}
  \item Did you mention the license of the assets?
    \answerNA{}
  \item Did you include any new assets either in the supplemental material or as a URL?
    \answerNA{}
  \item Did you discuss whether and how consent was obtained from people whose data you're using/curating?
    \answerNA{}
  \item Did you discuss whether the data you are using/curating contains personally identifiable information or offensive content?
    \answerNA{}
\end{enumerate}

\item If you used crowdsourcing or conducted research with human subjects...
\begin{enumerate}
  \item Did you include the full text of instructions given to participants and screenshots, if applicable?
    \answerNA{}
  \item Did you describe any potential participant risks, with links to Institutional Review Board (IRB) approvals, if applicable?
    \answerNA{}
  \item Did you include the estimated hourly wage paid to participants and the total amount spent on participant compensation?
    \answerNA{}
\end{enumerate}

\end{enumerate}




\end{document}